\documentclass[12pt]{article}
\usepackage{setspace}
\usepackage{amssymb} 
\usepackage{amsmath} 
\usepackage{amsthm}
\usepackage{enumerate}
\usepackage{array}
\usepackage{multirow}
\usepackage{float}
\usepackage{caption}
\usepackage{subcaption}
\usepackage{natbib}
\usepackage{graphicx}
\usepackage[a4paper,left=1in,right=1in,top=1in,bottom=1in]{geometry}
\PassOptionsToPackage{normalem}{ulem}
\usepackage{ulem}
\usepackage{xcolor}
\usepackage[
    colorlinks,
    linkcolor={blue!50!black},
    citecolor={blue!50!black},
    urlcolor={blue!80!black}]{hyperref}

\theoremstyle{plain}

\providecommand{\theoremname}{Theorem}

\begin{document}
\sloppy

\begin{singlespace}

\title{Co-data Learning for Bayesian Additive Regression Trees}

 \author{\small{Jeroen M. Goedhart\footnote{Corresponding author, E-mail address: j.m.goedhart@amsterdamumc.nl} \textsuperscript{a}, Thomas Klausch\textsuperscript{a}, Jurriaan Janssen\textsuperscript{b}, Mark A. van de Wiel\textsuperscript{a}}}
\date{}
\maketitle
\noindent\textsuperscript{a}\footnotesize{Department of Epidemiology and Data Science, Amsterdam Public Health Research Institute,
Amsterdam University Medical Centers Location AMC, Meibergdreef 9, the Netherlands}\\

\noindent\textsuperscript{b}\footnotesize{Department of Pathology, Cancer Center Amsterdam,
Amsterdam University Medical Centers Location VUMC, De Boelelaan 1117, the Netherlands}

\begin{abstract}
Medical prediction applications often need to deal with small sample sizes compared to the number of covariates. Such data pose problems for prediction and variable selection, especially when the covariate-response relationship is complicated. To address these challenges, we propose to incorporate co-data, i.e. external information on the covariates, into Bayesian additive regression trees (BART), a sum-of-trees prediction model that utilizes priors on the tree parameters to prevent overfitting. To incorporate co-data, an empirical Bayes (EB) framework is developed that estimates, assisted by a co-data model, prior covariate weights in the BART model. The proposed method can handle multiple types of co-data simultaneously. Furthermore, the proposed EB framework enables the estimation of the other hyperparameters of BART as well, rendering an appealing alternative to cross-validation. We show that the method finds relevant covariates and that it improves prediction compared to default BART in simulations. If the covariate-response relationship is nonlinear, the method benefits from the flexibility of BART to outperform regression-based co-data learners. Finally, the use of co-data enhances prediction in an application to diffuse large B-cell lymphoma prognosis based on clinical covariates, gene mutations, DNA translocations, and DNA copy number data.

\end{abstract}
\end{singlespace}
\newpage

\section{Introduction}
\label{sec:intro}
Modern prediction models that deal with many covariates often make
use of an additional structure in the covariate-response relationship
to improve predictions and variable selection. A widely-used structure
is sparsity, employed in lasso regression (\cite{LassoRegression1996})
and the horseshoe (\cite{Horseshoe}). Another method to embed a structure
is by employing an \textit{a priori} weighting scheme for covariates.
For instance, in the grouped lasso (\cite{GroupedLasso2006}), such
scheme corresponds to the amount of regularization each group of covariates
obtains, and for tree-based methods, the scheme corresponds to the
prior probabilities that covariates are selected in the splitting
rules.

Known structures can be embedded when prior knowledge on the application
at hand is available. For example, when few covariates are \textit{a
priori} expected to have an effect, a sparsity structure may be assumed.
Alternatively, one may learn the structure using external data sources.
External data may be discrete such as a grouping of the covariates
and, in the context of Omics, a known gene signature, or this data
may be continuous such as estimated p-values and effect size estimates
of the covariates from a previous related study. We consider scenarios
where such external data, from now on termed co-data (\cite{CoData}),
is available. 

Several linear models learn a structure in the covariate-response
relationship from co-data. Such models often utilize a Bayesian framework
to naturally incorporate the co-data through the priors, either by
employing empirical or full Bayes. Most of these learners only handle
discrete co-data, often in the form of an \textit{a priori} grouping
of the covariates. Examples include GRridge (\cite{GRidgevandeWiel}),
which employs empirical Bayes (EB), and Graper (\cite{Velten2021AdaptivePenalization}),
which is full Bayes. However, recently, linear models that allow multiple
sources of co-data, both continuous and discrete, were developed:
Ecpc (\cite{ecpc}), which is EB-based, and Fwen (\cite{FeatureWeightedElasticNet}),
which differs from the aforementioned methods by utilizing non-Bayesian
estimation to learn covariate weights.

For more flexible predictions models such as tree-based models, fewer
co-data methods are available, whereas these models may greatly benefit
from an additional structure to reduce the search in parameter space.
A notable contribution along these lines is CoRF, which learns covariate
weights from both discrete and continuous co-data to enrich a random
forest model (\cite{teBeest2017codata}). Here, we develop an EB-based
method to incorporate discrete and continuous co-data into Bayesian
additive regression trees (BART) (\cite{chipmanBART2010}).

BART is a sum-of-trees prediction model, embedded in a Bayesian framework,
which may estimate continuous and binary (\cite{chipmanBART2010}),
survival (\cite{surivalBART2016}), and count responses (\cite{BARTWAIC}).
The main advantage of a sum-of-trees model is that both additive and
interaction effects are captured and that these effects are learned
nonparametrically from the primary data. To reduce the flexibility
of the sum-of-trees, BART utilizes regularization priors on the tree
parameters to prevent overfitting by favoring shallow trees and small
response estimates. BART distinguishes itself from random forest by
mainly modeling low-order interaction effects, whereas a random forest
by default grows deeper trees and hence incorporates high-order interaction
effects. Therefore, BART is more interpretable and it lends itself
better for variable selection compared to a random forest.

BART has already been extended to embed additional structures in the
covariate-response relationship. DART, proposed by \citet{Linero2018barthighdimensional},
incorporates a sparse structure, while OG-BART incorporates a discrete,
possibly overlapping, grouping structure with an additional sparsity
layer (\cite{GroupingBART}). Both methods employ full Bayes to estimate
the structure. A disadvantage is that those methods are specifically
designed for sparse settings. However, high-dimensional applications
may often be rather dense, as argued for many omics settings (\cite{BOYLE2017}). 

We present EB-coBART, a BART model that learns from both the primary
data and co-data to embed a structure on the covariates in the form
of a weighting scheme. \citet{bleich2014variable} showed that a weighting scheme for the covariates based on
prior knowledge may lead to better predictions compared to default
BART. However, a fairly crude method was proposed by putting twice
as much weight on covariates that were known to be important. It
is in general nontrivial how to determine those weights. Furthermore,
it is difficult to judge the informativeness of the prior knowledge
for the data set at hand. EB-coBART has a more objective and data-driven
approach by learning the weighting scheme by Empirical Bayes combined
with a parsimonious co-data model. The parsimony protects against
overfitting and ensures stable estimated weights for the covariates. 

Our method has several assets compared to aforementioned approaches
to co-data learning. First, EB-coBART handles both discrete and continuous
co-data. Second, the weighting scheme of the covariates is learned
by both the primary and the co-data, which protects better against
overfitting compared to approaches that learn the weights only from
the primary data. Third, it does not require sparsity and hence many (small) effects may be modeled. Fourth, co-data is naturally incorporated using Empirical Bayes (EB), in contrast to CoRF which relies on a more ad hoc incorporation of the co-data
(\cite{teBeest2017codata}). Furthermore, EB is easily implemented in
existing BART software, whereas full Bayes approaches require rebuilding
the algorithm from scratch.

In addition, our EB framework may also be used to estimate the other
hyperparameters of BART, thereby avoiding cross-validation. While
we focus on estimating co-data driven hyperparameters, we also provide
EB estimators for the other hyperparameters and showcase them in an
application.

The remainder of this article is organized as follows. We start by
reviewing BART in \autoref{sec:Summary-of-BART}. In \autoref{sec:EB-coBART},
we first derive a general empirical Bayes framework that may estimate
any hyperparameter of BART. Then, we describe how the co-data model
is included to this framework to estimate the covariate weights. Next,
we present simulations in \autoref{sec:Simulations} and we illustrate
our method in an application to lymphoma patients in \autoref{sec:Applications}.
We conclude with a summary and a discussion in \autoref{sec:Discussion}

\section{Summary of BART}
\label{sec:Summary-of-BART}

Bayesian additive regression trees (BART), proposed by \citet{chipmanBART2010}, is a Bayesian sum-of-trees model. The
main idea is to sum several weak learners, i.e. the individual trees,
to form a powerful prediction model. Individual trees are discouraged
to have a large effect on the sum by employing regularization priors
that favor shallow trees and small terminal node values. Because our
proposed method relies on several aspects of the original BART model,
we present a detailed summary. We focus on regression, but BART and
our proposed method also generalize to classification using the data
augmentation method of \citet{albert1993bayesian}. 

\subsection{A sum of trees model with regularization priors}

Suppose we have a data set $D=\left\{ \left(y_{i},\boldsymbol{x}_{i}\right)\right\} _{i=1}^{N}$
consisting of $N$ observations of a normally distributed response
$y_{i}$ and a $p$-dimensional vector of covariates $\boldsymbol{x}_{i}=\left(x_{1},...,x_{p}\right)$
with $x_{j}$ representing covariate $j.$ In prediction, we then
model $y_{i}$ by $y_{i}=f\left(\boldsymbol{x}_{i}\right)+\epsilon_{i},\quad\epsilon_{i}\sim\mathcal{N}\left(0,\sigma^{2}\right),$
and we aim to estimate the function $f\left(\boldsymbol{x}_{i}\right).$

BART approximates $f\left(\boldsymbol{x}_{i}\right)$ by a sum-of-trees
model $G\left(\boldsymbol{x}_{i};\boldsymbol{\mathcal{T}},\boldsymbol{\mathcal{M}}\right)$
with input $\boldsymbol{x}_{i}$ and parameters $\boldsymbol{\mathcal{T}}$
and $\boldsymbol{\mathcal{M}}$: 
\begin{equation}
f\left(\boldsymbol{x}_{i}\right)\approx G\left(\boldsymbol{x}_{i};\mathcal{\boldsymbol{\mathcal{T}}},\boldsymbol{\mathcal{M}}\right)\equiv\sum_{t=1}^{K}g_{t}\left(\boldsymbol{x}_{i};\mathcal{T}_{t},\mathcal{M}_{t}\right).\label{sum of trees model}
\end{equation}
Here, $g_{t}$ denotes the $t$th regression tree, having tree structure
parameter $\mathcal{T}_{t}$ and terminal node parameter $\mathcal{M}_{t},$
and $K$ denotes the number of trees. Parameters $\boldsymbol{\mathcal{T}}$
and $\boldsymbol{\mathcal{M}}$ collect all tree structures $\mathcal{T}_{t}$
and all terminal node parameters $\mathcal{M}_{t},$ respectively. 

In BART, the sum-of-trees model is embedded in a Bayesian framework
with a Gaussian likelihood
\begin{equation}
\pi\left(\boldsymbol{y}\mid\boldsymbol{X},\mathcal{\boldsymbol{\mathcal{T}}},\boldsymbol{\mathcal{M}},\sigma^{2}\right)=\prod_{i=1}^{N}\mathcal{N}\left(y_{i};G\left(\boldsymbol{x}_{i};\mathcal{\boldsymbol{\mathcal{T}}},\boldsymbol{\mathcal{M}}\right),\sigma^{2}\right),\label{eq:likelihood of BART}
\end{equation}
with $\boldsymbol{y}=\left(y_{1},...,y_{N}\right)^{T}$ and\textbf{
$\boldsymbol{X}\boldsymbol{=}\left(\boldsymbol{x}_{1}^{T},...,\boldsymbol{x}_{N}^{T}\right)^{T},$}
and a regularization prior distribution $\pi\left(\boldsymbol{\mathcal{T}},\mathcal{\boldsymbol{M}}\right)$.
For the error variance $\sigma^{2},$ the standard prior $\pi\left(\sigma^{2}\right)=\mathcal{IG}\left(\sigma^{2};\frac{\nu}{2},\frac{\nu\lambda}{2}\right),$
with hyperparameters $\nu$ and $\lambda,$ is employed.

\subsection{Prior specification of the tree parameters}

The prior $\pi\left(\boldsymbol{\mathcal{T}},\mathcal{\boldsymbol{M}}\right)$
on the tree parameters is simplified by assuming that each tree $t$
is independently and identically distributed with prior $\pi\left(\mathcal{T}_{t},\mathcal{M}_{t}\right)=\pi\left(\mathcal{M}_{t}\mid\mathcal{T}_{t}\right)\pi\left(\mathcal{T}_{t}\right):$
\begin{equation}
\pi\left(\boldsymbol{\mathcal{T}},\boldsymbol{\mathcal{M}}\right)=\prod_{t=1}^{K}\pi\left(\mathcal{M}_{t}\mid\mathcal{T}_{t}\right)\pi\left(\mathcal{T}_{t}\right),\label{eq:Full tree prior}
\end{equation}
 with tree structure prior $\pi\left(\mathcal{T}_{t}\right)$ and
terminal node prior $\pi\left(\mathcal{M}_{t}\mid\mathcal{T}_{t}\right).$ 

Priors $\pi\left(\mathcal{T}_{t}\right)$ and $\pi\left(\mathcal{M}_{t}\mid\mathcal{T}_{t}\right)$
are set such that each tree is expected to only contribute slightly
to the overall fit $f\left(\boldsymbol{x}\right)$. This regularization
is achieved by favoring shallow trees with small terminal node values.
Before specifying these priors, we review the, for us, relevant parameters
of a tree.

\subsubsection{Tree parameterization}

The tree structure $\mathcal{T}_{t}$ of tree $t$ is parameterized
by a set of nodes which are internal or terminal 
\begin{equation}
\mathcal{T}_{t}=\left(\xi_{1t}(x_{j},a_{j},d),\ldots,\xi_{Z_{t}t}(x_{j},a_{j},d),\omega_{1t}(d),\ldots,\omega_{L_{t}t}(d)\right),
\end{equation}
with $z=1,\ldots,Z_{t}$ indexing the internal nodes $\xi_{zt},$
and $l=1,\ldots,L_{t}$ indexing the terminal nodes $\omega_{lt}.$
An internal node $\xi$ is parameterized by a binary splitting rule
$\left\{ x_{j}\leq a_{j}\right\} ,$ with $x_{j}$ the chosen splitting
variable and $a_{j}$ a splitting value within the range of covariate
$j.$ Both the internal nodes and the terminal nodes have a depth
parameter $d,$ with the root node having $d=0.$ Index variables
$z$ and $l$ move from top to bottom and from left to right. We omitted
the topological structure of the tree.

Terminal node $\omega_{lt}$ also has a parameter $\mu_{lt}$, which
represents the response estimate for the given node. All $L_{t}$
estimates are collected in the terminal node parameter $\mathcal{M}_{t}=\left(\mu_{1t},\ldots,\mu_{L_{t}t}\right)$
for tree $t.$

\subsubsection{Tree structure prior}

The tree structure prior $\mathcal{T}_{t}$ is chosen such that nodes
have a probability of $\alpha\left(1+d\right)^{-\beta}$ to be internal
and $1-\alpha\left(1+d\right)^{-\beta}$ to be terminal, with hyperparameters
$\alpha\in\left(0,1\right)$ and $\beta>0.$ Hyperparameters $\alpha$
and $\beta$ are usually set at $\alpha=0.95$ and $\beta=2.$ Hence,
shallow trees are \textit{a priori} favored because it becomes more
likely that a node is terminal for a larger depth $d$. 

For internal nodes, splitting variables need to be assigned. Splitting
variables $x_{j}$ are chosen from a categorical prior with prespecified
probabilities $\boldsymbol{S}$ for each $j$: $\boldsymbol{S}=\left(s_{1},\ldots,s_{p}\right).$
Default BART sets equal covariate weights: $s_{j}=1/p.$
In our proposed method, $\boldsymbol{S}$ will be estimated using
Empirical Bayes and co-data. 

The above specification leads to the following prior on tree structure
$\mathcal{T}_{t}:$
\begin{equation}
\pi_{\alpha,\beta,\boldsymbol{S}}\left(\mathcal{T}_{t}\right)\propto\left[\prod_{z=1}^{Z_{t}}\textrm{Categorical}\left(x_{jzt};\boldsymbol{S}\right)\right]\left[\prod_{z=1}^{Z_{t}}\alpha\left(1+d_{zt}\right)^{-\beta}\right]\left[\prod_{l=1}^{L_{t}}1-\alpha\left(1+d_{lt}\right)^{-\beta}\right],\label{eq:TreeStructue Tj}
\end{equation}
with $Z_{t}$ the number of internal nodes of tree $t$, $x_{jzt}$
covariate $j$ occurring in the splitting rule of the $z$th internal
node of tree $t,$ $L_{t}$ the number of terminal nodes of tree $t$,
and $d_{zt}$ and $d_{lt}$ the depths of the $z$th internal node
and the $l$th terminal node of tree $t$, respectively. The subscript
denotes the dependence of the prior on hyperparameters $\alpha,$
$\beta,$ and $\boldsymbol{S}.$ We suppressed the dependence of the
prior on the splitting values $a_{jz}$ because these are not relevant
for the purpose of this paper.

\subsubsection{Terminal node prior}

The $L_{t}$ terminal node parameters $\mu_{lt},$ collected in $\mathcal{M}_{t},$
of tree $t$ are \textit{a priori} assumed to be independent and identically
distributed with a centered normal distribution:
\begin{equation}
\pi_{k}\left(\mathcal{M}_{t}\mid\mathcal{T}_{t}\right)=\prod_{l=1}^{L_{t}}\mathcal{N}\left(\mu_{lt};0,\sigma_{\mu}^{2}\right),\qquad \sigma_{\mu}=\frac{0.5}k\sqrt{K},\label{eq:LeafNode prior}
\end{equation}
with hyperparameter $k$ typically $1\leq k\leq3,$ This prior favors
small terminal node values and hence reduces the effect of a single
tree on the overall fit. This reduction is more prominent for larger
$k$ or number of trees $K.$ 

\subsection{Draws from the posterior}

By employing the regularizing priors, given by \eqref{eq:Full tree prior},
\eqref{eq:TreeStructue Tj}, and \eqref{eq:LeafNode prior}, and
the likelihood, i.e. \eqref{eq:likelihood of BART}, a posterior
of the model parameters $\left(\mathcal{\boldsymbol{\mathcal{T}}},\boldsymbol{\mathcal{M}},\sigma^{2}\right)$
may be deduced. Because of the large discrete tree space, this posterior
is analytically intractable, and hence a Gibbs sampler is employed.
This sampling algorithm is described elsewhere (\cite{tan2019bayesiantutorial,KapelnerBART}).
In Supplementary Section 1, we provide a short summary
of this algorithm using our notation. 

The Gibbs samples represent, after a suitable burn-in period, posterior
samples of the sum-of-trees $\sum_{t=1}^{K}g_{t}\left(\boldsymbol{x}_{i};\mathcal{T}_{t},\mathcal{M}_{t}\right)$,
which then allow inference on the response $Y.$ We assess convergence
of the MCMC chain by estimating the Gelman-Rubin diagnostic (\cite{GelmanRubinDiagnostic})
for the error variance $\sigma^{2}$ for continuous responses and
for $\sum_{t=1}^{K}g_{t}\left(\boldsymbol{x}_{i};\mathcal{T}_{t},\mathcal{M}_{t}\right)$
for binary responses, as proposed by \citet{BARTRPackage}.

\section{EB-coBART}
\label{sec:EB-coBART}
It may be beneficial for BART to upweight certain (groups of) covariates
that are expected to be informative for the response $Y.$ In BART,
this is naturally done by modifying hyperparameter $\boldsymbol{S}=\left(s_{1},\ldots,s_{p}\right)$
of the categorical splitting variable prior, which normally defaults
to a discrete uniform: $s_{j}=1/p$. Our main contribution is to impose a weighting
structure in $\boldsymbol{S}$ in a data-driven manner. To do so,
we combine empirical Bayes (EB) with a co-data model.

Empirical Bayes allows adaptively learning the structure in $\boldsymbol{S}$
with a central role for the primary data $D=\left\{ \left(y_{i},\boldsymbol{x}_{i}\right)\right\} _{i=1}^{N}.$
However, estimating $\boldsymbol{S}$ by pure EB requires estimating
a $p$-dimensional hyperparameter, which likely leads to overfitting.
Hence, we model the EB estimate of $\boldsymbol{S}$ by a co-data
model, which renders a substantial reduction in dimension of the estimated
hyperparameter. Additionally, the co-data model ensures a natural
incorporation of external information on the covariates into BART.

We start by deriving EB for BART with the primary goal to estimate
$\boldsymbol{S}$. In addition, we show EB estimates of the other
hyperparameters $\left(\alpha,\beta,k,\nu,\lambda\right)$, which
are normally estimated by cross-validation. Next, we describe our
co-data model that guides the EB-estimate of $\boldsymbol{S}$. This
completes our proposed methodology, which we call EB-coBART. We end
this section with some remarks on choices for the other hyperparameters.

\subsection{Empirical Bayes for BART}
\label{subsec:Empirical Bayes for BART}

We collect all hyperparameters of BART in the vector $\boldsymbol{\rho}=\left(\alpha,\beta,k,\nu,\lambda,\boldsymbol{S}\right),$
which we then estimate by $\hat{\boldsymbol{\rho}}$ using empirical
Bayes (EB). To do so, we maximize the marginal likelihood of BART
w.r.t. $\boldsymbol{\rho}$: 
\begin{equation}
\hat{\boldsymbol{\rho}}=\arg\max_{\boldsymbol{\rho}}\;m_{\boldsymbol{\rho}}\left(\boldsymbol{y}\mid\boldsymbol{X}\right),\label{eq:maximization of marglik}
\end{equation}
with $m_{\boldsymbol{\rho}}\left(\boldsymbol{y}\mid\boldsymbol{X}\right)$
the marginal likelihood of BART and subscript $\boldsymbol{\rho}$
denoting dependence on the hyperparameters. 

\citet{BoatmanMargLikBart} showed that, under a reparameterization
of the terminal node prior, the marginal likelihood requires a summation
over the full tree space. This summation is exponentially large and
prevents direct maximization of the marginal likelihood. We therefore
rely on an approximation algorithm. 

\subsubsection{Monte Carlo EM algorithm}

To approximate \eqref{eq:maximization of marglik}, we employ the
Monte Carlo EM algorithm derived by \citet{Casella2001empirical}.
In essence, this algorithm utilizes Gibbs samples to approximate the expectation
step of the EM algorithm applied to the marginal likelihood. For BART,
this method amounts to the following iterative algorithm:
\begin{align}
\boldsymbol{\hat{\rho}}^{(q+1)} & \approx\underset{\boldsymbol{\rho}}{\arg\max}\frac{1}{n_{mc}}\sum_{m=1}^{n_{mc}}\log\left[\pi\left(\boldsymbol{y}\mid\boldsymbol{X},\boldsymbol{\mathcal{T}}_{m}^{(q)},\boldsymbol{\mathcal{M}}_{m}^{(q)},\sigma_{m}^{2(q)}\right)\pi_{\boldsymbol{\rho}}\left(\boldsymbol{\mathcal{T}}_{m}^{(q)},\boldsymbol{\mathcal{M}}_{m}^{(q)},\sigma_{m}^{2(q)}\right)\right]\nonumber \\
 & =\underset{\boldsymbol{\rho}}{\arg\max}\sum_{m=1}^{n_{mc}}\log\left[\pi_{\boldsymbol{\rho}}\left(\boldsymbol{\mathcal{T}}_{m}^{(q)},\boldsymbol{\mathcal{M}}_{m}^{(q)},\sigma_{m}^{2(q)}\right)\right],\label{eq:General form EM algorith for BART-1}
\end{align}
with $m$ indexing the Gibbs samples, $n_{mc}$ the total number of
Gibbs samples, $q$ the iteration index, and $\left(\boldsymbol{\mathcal{T}}_{m}^{(q)},\boldsymbol{\mathcal{M}}_{m}^{(q)},\sigma_{m}^{2(q)}\right)$
the $m$th posterior sample of the model parameters of BART with the
posterior evaluated at $\boldsymbol{\rho}=\hat{\boldsymbol{\rho}}^{(q)}.$
Thus, $\pi_{\boldsymbol{\rho}}\left(\boldsymbol{\mathcal{T}}_{m}^{(q)},\boldsymbol{\mathcal{M}}_{m}^{(q)},\sigma_{m}^{2(q)}\right)=\pi_{\alpha,\beta,\boldsymbol{S},k}\left(\boldsymbol{\mathcal{T}}_{m}^{(q)},\boldsymbol{\mathcal{M}}_{m}^{(q)}\right)\pi_{\nu,\lambda}\left(\sigma_{m}^{2(q)}\mid\boldsymbol{\mathcal{T}}_{m}^{(q)},\boldsymbol{\mathcal{M}}_{m}^{(q)}\right)$
evaluates the prior probabilities of these posterior samples as a
function of the hyperparameters $\boldsymbol{\rho}.$ The prior probabilities
are evaluated using \eqref{eq:Full tree prior}, \eqref{eq:TreeStructue Tj},
\eqref{eq:LeafNode prior}, and $\pi_{\nu,\lambda}\left(\sigma_{m}^{2(q)}\mid\boldsymbol{\mathcal{T}}_{m}^{(q)},\boldsymbol{\mathcal{M}}_{m}^{(q)}\right)=\mathcal{IG}\left(\sigma_{m}^{2(q)};\frac{\nu}{2},\frac{\nu\lambda}{2}\right).$
A derivation of \eqref{eq:General form EM algorith for BART-1}
is given in Supplementary Section $2$ (eq. $2.5$).

Solving \eqref{eq:General form EM algorith for BART-1} then renders
the iterative estimates of the hyperparameters $\left(\alpha,\beta,k,\nu,\lambda,\boldsymbol{S}\right):$
\begin{align}
\left(\hat{\alpha}^{(q+1)},\hat{\beta}^{(q+1)}\right) & =\underset{\alpha,\beta}{\arg\max}\sum_{m=1}^{n_{mc}}\sum_{t=1}^{K}\left[\sum_{z=1}^{Z_{tm}^{(q)}}\log\left(\alpha\left(1+d_{ztm}^{(q)}\right)^{-\beta}\right)+\sum_{l=1}^{L_{tm}^{(q)}}\log\left(1-\alpha\left(1+d_{ltm}^{(q)}\right)^{-\beta}\right)\right],\label{eq:HyperparameterUpdatesTree} \\
\hat{k}^{(q+1)} & =\frac{2\sum_{m=1}^{n_{mc}}\sum_{t=1}^{K}L_{tm}^{(q)}}{\sqrt{\sum_{m=1}^{n_{mc}}\sum_{t=1}^{K}\sum_{l=1}^{L_{tm}^{(q)}}\left(\mu_{ltm}^{(q)}\right)^{2}}\sqrt{K}},\label{eq:HyperparameterUpdatesLeaf} \\
\left(\hat{\nu}^{(q+1)},\hat{\lambda}^{(q+1)}\right) & =\underset{\nu,\lambda}{\arg\max}\sum_{m=1}^{n_{mc}}\log\mathcal{IG}\left(\sigma_{m}^{2(q)};\frac{\nu}{2},\frac{\nu\lambda}{2}\right),\label{eq:HyperparameterUpdatesSigma} \\
\boldsymbol{\hat{S}}_{1}^{(q+1)} & =\left(b_{1}^{(q)}/B^{(q)},\ldots,b_{p}^{(q)}/B^{(q)}\right),\label{eq:HyperparameterUpdatesS}
\end{align}
with $b_{j}^{(q)}$ the sampled number of splitting rules with covariate
$j$ at iteration $q$, and $B^{(q)}$ the total number of sampled
splitting rules. At iteration $q,$ samples are taken from the posterior
of BART using hyperparameter $\hat{\boldsymbol{\rho}}^{(q)}.$ Derivations
of the EB-estimates are found in Supplementary Section $2$.

Equation \eqref{eq:HyperparameterUpdatesS} shows feature-specific
EB estimates $\boldsymbol{\hat{S}}_{1}^{(q+1)}$ of hyperparameter $\boldsymbol{S},$ which likely leads to overfitting
because $p-1$ hyperparameter estimates are required. To address this
issue, we propose to model the feature-specific estimates using co-data, rendering a new estimator $\boldsymbol{\hat{S}}_{2}^{(q+1)}$ of $\boldsymbol{S}.$ 

\subsection{Co-data model}
\label{subsec:Co-data model}

Suppose we have $\kappa$ co-data variables of which we have complete
measurements for each covariate $j$. We represent a grouping
co-data variable having $G$ groups by dummy coding indicating which group covariate $j$ belongs to. Covariates with missing co-data observations may be defined as a separate group to account for potential information in the missingness (\cite{ecpc}).

We collect the co-data
measurements of each $j$ in the vector $\boldsymbol{c}_{j}\in\mathbb{R^{\kappa}}$
and define the $p\times\kappa$ co-data model matrix $\boldsymbol{\mathcal{C}}=\left(\boldsymbol{c}_{1}^{T},\ldots,\boldsymbol{c}_{p}^{T}\right)^{T}.$
We then employ $\boldsymbol{\mathcal{C}}$ for modeling the feature-specific
EB estimates $\boldsymbol{\hat{S}}_{1}^{(q+1)},$ i.e. \eqref{eq:HyperparameterUpdatesS}.

To link $\boldsymbol{\hat{S}}_{1}^{(q+1)}$ to
co-data $\boldsymbol{\mathcal{C}},$ we model the counts $b_{j}^{(q)}.$
We start by noting that each of the in total $B^{(q)}$ splitting
rules may be regarded as a Bernoulli trial for any covariate $j$:
covariate $j$ occurs or does not occur in the given splitting rule.
Therefore, count $b_{j}^{(q)}$ is the outcome of a binomial trial:
$b_{j}^{(q)}\sim\textrm{Bin}\left(w_{j}^{(q)},B^{(q)}\right),$ with
unknown probability of success $w_{j}^{(q)}$ for covariate $j$
at iteration $q.$ We then model $w_{j}^{(q)}$ using $\boldsymbol{\mathcal{C}}$
by employing a logistic regression model:
\begin{equation}
w_{j}^{(q)}=\textrm{expit}\left(\boldsymbol{c}_{j}^{T}\boldsymbol{\eta}^{(q)}\right),\qquad\textrm{for}\quad j=1,.\ldots,p,\label{eq:co-data model}
\end{equation}
with $\textrm{expit}\left(x\right)=e^{x}/\left(e^{x}+1\right),$ and
\textbf{$\boldsymbol{{\eta}}^{(q)}\in\mathbb{R^{\kappa}}$ }a regression parameter
vector. 

We estimate $w_{j}^{(q)}$ by $\hat{w}_{j}^{(q)},$ which will be
the co-data moderated EB-estimates $\boldsymbol{\hat{S}}_{2}^{(q+1)}$ of hyperparameter $\boldsymbol{S}.$
To determine $\hat{w}_{j}^{(q)},$ we first estimate $\boldsymbol{\eta}^{(q)}:$
\begin{equation}
\hat{\boldsymbol{\eta}}^{(q)}=\arg\max_{\boldsymbol{\eta}^{(q)}}\;\sum_{j=1}^{p}\log\left[\textrm{Bin}\left(\textrm{expit}\left(\boldsymbol{c}_{j}^{T}\boldsymbol{\eta}^{(q)}\right),B^{(q)}\right)\right],\label{eq:MaxLik co-data model}
\end{equation}
i.e. maximum likelihood maximization for logistic regression with
covariates $j$ serving as samples. Equation \eqref{eq:MaxLik co-data model}
ignores two dependencies between samples $j$. First, only one covariate
can be used for a given splitting rule. Hence, success in a given
Bernoulli trial for variable $j$ determines failure for the other
variables. The second dependency originates from the dependence between
the covariates, which induces dependencies between the splitting rules.
Because we are only interested in the point estimates $\hat{\boldsymbol{\eta}}^{(q)}$
and because the estimator \eqref{eq:MaxLik co-data model} is
consistent, we ignore these dependencies.

Estimate $\hat{\boldsymbol{\eta}}^{(q)}$ then determines the estimates
$\hat{w}_{j}^{(q)}=\textrm{expit}\left(\boldsymbol{c}_{j}^{T}\hat{\boldsymbol{\eta}}^{(q)}\right),$ which we collect in the co-data guided empirical Bayes estimator $\boldsymbol{\hat{S}}_{2}^{(q+1)}$ of $\boldsymbol{S}$ at iteration $q+1$:
\begin{equation}
\boldsymbol{\hat{S}}_{2}^{(q+1)}=\left(\hat{w}_{1}^{(q)},\ldots,\hat{w}_{p}^{(q)}\right).\label{eq:co-data EB estimates of S}
\end{equation}
Note that $\hat{\boldsymbol{\eta}}^{(q)}$ fully determines \eqref{eq:co-data EB estimates of S} given the co-data $\boldsymbol{\mathcal{C}}.$ Hence, we only estimate $\kappa$ hyperparameters instead of $p-1$ (which is the case for \eqref{eq:HyperparameterUpdatesS}) to determine $\boldsymbol{\hat{S}}_{2}^{(q+1)}$.  We therefore opt for a parsimonious model, i.e. \eqref{eq:co-data model}, with $\kappa\ll p$ and only linear effects. 

Equation \eqref{eq:co-data model} ensures that the co-data and the
primary data should be in agreement. If, for example, the feature
specific EB estimates of \eqref{eq:HyperparameterUpdatesS}
show a strong preference for certain covariates, while this preference
is not present in the co-data, the final up/down weighting of the
covariates will be diminished. Vice versa, if the co-data shows a
strong preference for a certain covariate, e.g. by a small p-value,
but BART does not recognize this covariate, the covariate will not
be substantially upweighted.

\subsection{Convergence of EB-coBART}
\label{subsec:Convergence of EB-coBART}

Equation \eqref{eq:co-data EB estimates of S} provides iterative updates
of hyperparameter $\boldsymbol{S}$ and hence a convergence criterion
is required. Typically, either the marginal likelihood or the estimated
hyperparameters, in this case $\hat{\boldsymbol{\eta},}$ are tracked
until they stabilize within a given tolerance level (\cite{Casella2001empirical}).
However, for BART, we found that both options are not feasible. 

First, tracking the marginal likelihood is difficult because its estimation
within reasonable computational time is nontrivial. Known sampling
methods like the harmonic mean estimator (\cite{HarmonicMeanEstimator})
or averaging the likelihood from a sample of the prior do not work
because too many samples are required to cover the tree space. \citet{BoatmanMargLikBart} estimated the marginal likelihood using prior
sampling, but only two covariates were considered.
For larger $p,$ a scenario we are interested in, the tree space becomes
too large for this method to work. Another option for marginal likelihood
estimation was proposed by \citet{ChibMargLik}. This option is
also unfeasible, because it requires an analytic expression of the
full conditional of the tree structure parameter $\mathcal{\boldsymbol{\mathcal{T}}}$
of BART, which is not available. Second, we empirically found that
tracking hyperparameter $\boldsymbol{\eta}$ until convergence fails
because of overfitting (Supplementary Section 3).

We therefore rely on the widely applicable information criterion (WAIC)
(\cite{watanabe2013widely}):
\begin{equation}
\textrm{WAIC}=-2\sum_{i=1}^{N}\log\left(E\left[\pi\left(y_{i}|\boldsymbol{\mathcal{T}},\boldsymbol{\mathcal{M}},\sigma^{2},\boldsymbol{x_{i}}\right)\right]\right)+2\sum_{i=1}^{N}\textrm{Var}\left[\log\left(\pi\left(y_{i}|\boldsymbol{\mathcal{T}},\boldsymbol{\mathcal{M}},\sigma^{2},\boldsymbol{x_{i}}\right)\right)\right],\label{eq:WAIC}
\end{equation}
with $\pi\left(y_{i}|\boldsymbol{\mathcal{T}},\boldsymbol{\mathcal{M}},\sigma^{2},\boldsymbol{x_{i}}\right)$
the predictive density, and the expectation and variance taken w.r.t.
the posterior of BART. We then halt the iterative hyperparameter updates
$\boldsymbol{\hat{S}}_{2}^{(q+1)}$ when the WAIC is at minimum. 

The WAIC as convergence criterion has some desirable properties. First,
the WAIC is easily computed for nonparametric methods such as BART, contrary to other information criteria. Second, the
WAIC is asymptotically equivalent to leave-one-out cross-validation
of the likelihood. Third, computationally, the WAIC is an efficient
stopping criterion because it may be estimated from the already available
posterior samples of BART (\cite{BARTWAIC}). 

As an alternative to the WAIC, the computationally more demanding
cross-validation (CV) may be used. In Supplementary Section 4,
we illustrate in a simulation that CV as stopping criterion leads
to similar predictive performance and variable selection results compared
to WAIC.

\subsection{Choices for the other hyperparameters}
\label{subsec:Choices-for-the}

Our derived EB scheme for BART (\eqref{eq:HyperparameterUpdatesTree}, \eqref{eq:HyperparameterUpdatesLeaf}, \eqref{eq:HyperparameterUpdatesSigma}, and \eqref{eq:HyperparameterUpdatesS})
may also be used to estimate the other hyperparameters $\left(\alpha,\beta,k,\nu,\lambda\right)$
of BART. Using EB for hyperparameter estimation may be an appealing
alternative to cross-validation, which relies on a subjective grid
of hyperparameter choices.

Estimation of multiple (correlated) hyperparameters, however, is intrinsically difficult. For example, for elastic net, a substantially less flexible prediction model compared to BART, it was proven that joint estimation of two hyperparameters already causes identifiability issues (\cite{ElasticNetMargLikMirrelijn}). Also, automatic hyperparameter tuning does not directly account for model interpretability.

Therefore, our default choice is to fix the tree structure hyperparameters $\alpha$ and $\beta$ at two flexibility levels: rigid ($\alpha=0.1,$ $\beta=4$), favoring shallow trees and thus interpretability, and flexible ($\alpha=0.95,$ $\beta=2$), favoring less shallow trees. We stress that the flexible model corresponds to the default settings of BART (\cite{chipmanBART2010}). Hyperparameter $k$ is fixed at $k=1$ for the rigid model and $k=2$ for the flexible model, acknowledging that  shallow trees have more observations in the terminal nodes and thus require less shrinkage of the terminal node parameters. For the error variance hyperparameters $\left(\nu,\lambda\right),$ we employ the default settings proposed by \citet{chipmanBART2010}.

However, for high-dimensional data, the predictions of BART may improve by hyperparameter fine-tuning (\cite{chipmanBART2010}). In the high-dimensional application (\autoref{sec:Applications}), we illustrate that hyperparamater estimation of $\alpha$ and $k$ using the derived EB-estimator in \eqref{eq:HyperparameterUpdatesTree} and \eqref{eq:HyperparameterUpdatesLeaf} renders a slight improvement in predictive performance compared to fixing $\alpha$ and $k.$

\section{Simulations}
\label{sec:Simulations}

We compare EB-coBART to BART when informative co-data is available in a high-dimensional data setting. To do so, we consider two functions $f$ that specify a covariate-response relationship: a sparse and nonlinear function $f_{SN}$ (\autoref{subsec:Sparse-and-nonlinear}),
and a dense and linear function $f_{LD}$ (\autoref{subsec:Dense-and-linear}). For $f_{SN}$, we employ discrete co-data in the form of a grouping
structure, while for $f_{LD}$, we employ continuous co-data in the
form of a noisy version of the true linear effect sizes. This set-up ensures that we evaluate the influence on BART of different types of co-data. Note that OG-BART
(\cite{GroupingBART}) can handle discrete co-data as well, but because
its implementation is lacking in the public domain, we do not perform
a comparison with this method. \autoref{subsec:Uninformative-co-data}
deals with uninformative grouping co-data to evaluate whether EB-coBART
recognizes such co-data by not upweighting certain groups.

For both functions, we consider multiple simulation settings which
are specified in the subsections. In each setting and for each function,
we simulate $N_{sim}=500$ data sets. For each data set, we first
fit BART, i.e. BART with equal covariate weights $s_{j}=1/p.$ Then, we
iteratively update $s_{j}$ according
to estimator \eqref{eq:co-data EB estimates of S} until the WAIC, \eqref{eq:WAIC}, is at minimum. BART fitted with these estimated
covariate weights then corresponds to our method EB-coBART. To fit
BART, we employ the R package dbarts (\cite{dbarts}),
and to estimate the WAIC, we employ the R package loo (\cite{PackageLOO}).

We consider a rigid tree model ($\alpha=0.1,$ $\beta=4,$ $k=1$),
and a flexible tree model ($\alpha=0.95,$ $\beta=2,$ $k=2$), as
explained in \autoref{subsec:Choices-for-the}. Thus, for each
data set, we have four final BART fits: rigid BART, flexible
BART, which is the default, rigid EB-coBART, and flexible EB-coBART.

We set the other hyperparameters settings of all
BART models as follows. We fix the error variance hyperparameters:
$\nu=10$ and $\lambda$ such that the $75$\% quantile of the prior
equals $2/3\hat{\textrm{Var}}\left(\boldsymbol{y}\right),$ with $\hat{\textrm{Var}}\left(\boldsymbol{y}\right)$
the estimated variance of the simulated response $\boldsymbol{y}.$
We also fix the number of trees to $K=50$ to balance prediction and
variable selection (\cite{chipmanBART2010,bleich2014variable}). 

For each simulated data set, we estimate the predictive performance
on a large ($N_{test}=500$) independent test set for both BART and EB-coBART. We quantify the performance by the prediction
mean square error (PMSE), i.e. $\textrm{PMSE}=N^{-1}\sum_{i=1}^{N}\left(y_{i}-\hat{y}_{i}\right)^{2}$
and $\hat{y}_{i}$ denoting the prediction for sample $i.$ Predictions
$\hat{y}_{i}$ are obtained by averaging posterior samples of the sum-of-trees (\cite{chipmanBART2010}). We collect variable importance results by monitoring the co-data-moderated EB estimates $\hat{w}_{j}^{(q)},$ \eqref{eq:co-data EB estimates of S},
which relate directly to how often the covariates occur in the tree
ensemble.

\subsection{Sparse and nonlinear setting} 
\label{subsec:Sparse-and-nonlinear}

Response $y_{i}$ is generated by $y_{i}=f_{SN}\left(\boldsymbol{x}_{i}\right)+\epsilon_{i},$
with noise $\epsilon_{i}\sim\mathcal{N}\left(0,1\right),$ for $i=1,\ldots,N,$
and with
\begin{equation}
f_{SN}\left(\boldsymbol{x}_{i}\right)=10\sin\left(\pi x_{i1}x_{i2}\right)+10x_{i3}+20\left(x_{i101}-0.5\right)^{2}+10x_{i102},\label{eq:sparse nonlinear function}
\end{equation}
and $x_{ij}\stackrel{i.i.d.}{\sim}\textrm{Unif}\left(0,1\right),$
for $j=1,\dots,p,$ and $p=500.$ Thus, covariates $\left\{ 1,2,3,101,102\right\} $
are predictive for the response and the remaining $495$ covariates
are noise. 

Co-data are defined as a grouping structure with $G=\{5,20\}$ groups.
We set equal-sized groups of size $100$ for $G=5$ and size $25$
for $G=20.$ We then assign covariates $1,2,\ldots,p/G$ to group
$1$, covariates $p/G+1,\ldots,2p/G$ to group $2$, et cetera. Group
$G$ consists of covariates $j=\left[\left(G-1\right)p/G+1\right],\ldots,p$. This distribution of covariates among the groups ensures that predictive
covariates $\{1,2,3\}$ are always in the same group (Group $1$ for
$G=5$ and $G=20$) and that predictive covariates $\{101,102\}$
are always in the same group (Group $2$ for $G=5$ and Group $5$
for $G=20$). For both $G$'s, we consider two samples sizes ($N=100$
and $N=200$).

\begin{figure}[h!]
\begin{centering}
\subfloat[\label{fig:GroupWeights_G5}]{\centering\includegraphics[width=\textwidth]{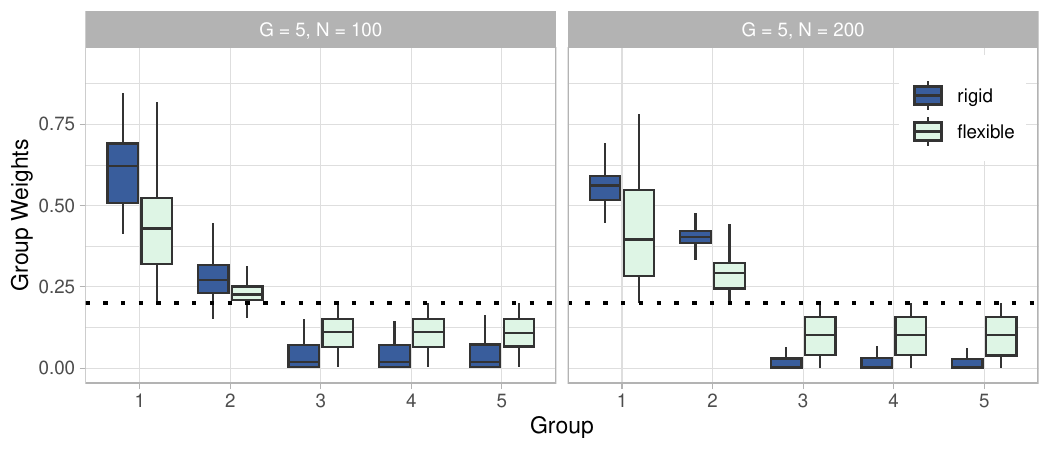}}

\subfloat[\label{fig:GroupWeights_G20}]{\centering\includegraphics[width=\textwidth]{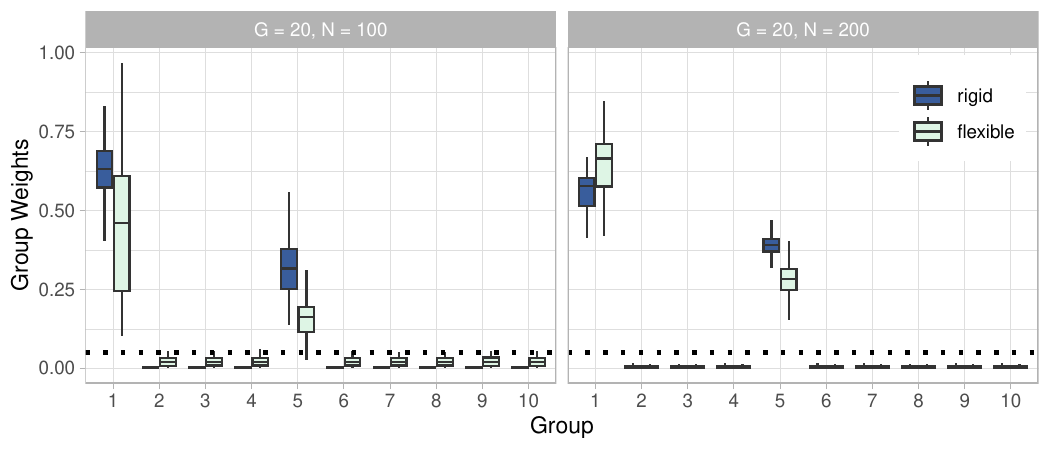}}\caption{\small{Boxplots of the co-data moderated EB estimates of the covariate weights of EB-coBART across the $500$
simulated data sets for different simulation settings. Figure \textbf{(a)} shows results for $G=5$ and Figure \textbf{(b)} for $G=20,$ for which we only depict the first $10$ groups for visualization. For each group, left (blue) boxplot corresponds to the rigid tree setting ($\alpha=0.1,$ $\beta=4,$ $k=1$) and right (green) boxplot to the flexible one ($\alpha=0.95,$ $\beta=2,$ $k=2$). Outliers are not shown. The horizontal dotted lines correspond to equal group weights ($0.2$ for $G=5;$ $0.05$ for $G=20$).}}
\label{fig:GroupWeights_Friedman}
\end{centering}
\end{figure}

Our results demonstrate that EB-coBART upweights the predictive groups ($1$ and $2$ for $G=5;$ $1$ and $5$ for $G=20$) and downweights the non-predictive groups in all simulation settings (\autoref{fig:GroupWeights_Friedman}). This upweighting effect is stronger for the rigid tree models because shallow trees include fewer noisy, non-relevant covariates in this sparse setting. Increasing the sample size from $N=100$ to $N=200$ reduces the variability in the group-specific estimates across the data sets as expected.

\autoref{tab:Average-PMSE} depicts the average PMSE for the four BART models, and  three competitors: random forest, and two other co-data learners that can handle both discrete and continuous co-data: Ecpc (\cite{ecpc}) and CoRF (\cite{teBeest2017codata}). For Ecpc, we include a posthoc variable selection procedure based on the elastic net penalty (\cite{GRidgevandeWiel,ecpc}), using the optimal number of covariates $p_{\textrm{sel}}=5.$ Performances were worse
with $p_{\textrm{sel}}=\{10,20\}.$

\begin{table}[h]
\centering{}\caption{\small{Average PMSE across data sets for several simulation settings for BART and EB-coBART in the rigid tree setting ($\alpha=0.1,$ $\beta=4,$ $k=1$) and the flexible tree setting ($\alpha=0.95,$ $\beta=2,$ $k=2$). Also included are competitors: random forest, and co-data learners: CoRF and Ecpc with and without posthoc variable selection.}}
\label{tab:Average-PMSE}
\begin{tabular}{ccccc}
 & $N=100,$ & $N=100,$ & $N=200,$ & $N=200,$\tabularnewline
 & $G=5$ & $G=20$ & $G=5$ & $G=20$\tabularnewline
\hline 
\hline 
Flexible BART & $11.4$ & $11.4$ & $4.58$ & $4.58$\tabularnewline
Flexible EB-coBART & $10.1$ & $7.63$ & $\boldsymbol{4.23}$ & $\boldsymbol{3.23}$\tabularnewline
Rigid BART & $9.30$ & $9.30$ & $4.94$ & $4.94$\tabularnewline
Rigid EB-coBART & $\boldsymbol{8.81}$ & \textbf{$\boldsymbol{7.47}$} & $4.66$ & $4.27$\tabularnewline
Ecpc (no variable selection) & $24.6$ & $16.9$ & $17.9$ & $11.6$\tabularnewline
Ecpc (variable selection) & $11.7$ & $8.43$ & $8.67$ & $7.46$\tabularnewline
Random forest & $20.7$ & $20.7$ & $15.4$ & $15.4$\tabularnewline
CoRF & $19.2$ & $15.0$ & $14.3$ & $11.6$\tabularnewline
\end{tabular}
\end{table}

For all simulation settings,  flexible or rigid EB-coBART performs best (\autoref{tab:Average-PMSE}). Furthermore, both flexible and rigid EB-coBART have
a substantially lower average PMSE compared to the competitors. For most settings, rigid and flexible BART also outperform the competitors. The exception is for $N=100,$ $G=20$ where Ecpc with variable selection has a lower average PMSE than flexible BART. Rigid EB-coBART has a lower PMSE than flexible EB-coBART for $N=100$ and flexible EB-coBART has a lower PMSE for $N=200.$ In this sparse simulation setting, more regularization on the tree depth is required for smaller sample sizes. 

\begin{figure}[h!]
\begin{centering}
\includegraphics[bb=0bp 18bp 530bp 144bp,clip,width=\textwidth]{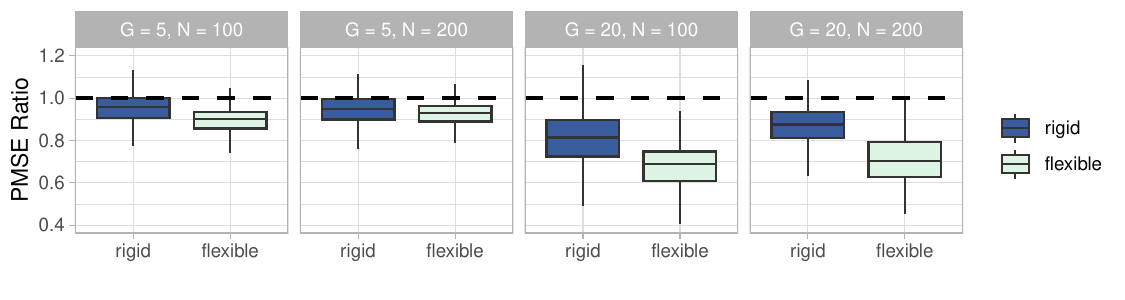}
\par\end{centering}
\centering{}\caption{\small{Boxplot of the ratio $\textrm{PMSE}_{\textrm{EBcoBART}}/\textrm{PMSE}_{\textrm{BART}}$
across the $500$ simulated data sets for both the rigid tree models
(blue, left) and the flexible tree models (green, right). The four panels correspond
to different simulation settings.}}
\label{fig:RatioPMSE}
\end{figure}

\autoref{fig:RatioPMSE} compares the relative PMSE of EB-coBART to that of BART demonstrating that EB-coBART has a smaller PMSE
than BART, i.e. $\textrm{PMSE}_{\textrm{EBcoBART}}/\textrm{PMSE}_{\textrm{BART}}<1,$
for most data sets in all considered simulation settings and for both
tree flexibility settings. Only for the rigid tree models in the $G=5$
setting, BART has a smaller PMSE than EB-coBART for a relatively
large percentage of simulated data sets ($19$\% for $N=100;$ $22$\%
for $N=200$).

Increasing the number of groups from $G=5$ to $G=20$ improves the
relative performance of EB-coBART compared to BART, because
the co-data has become relatively more informative.
A larger sample size ($N=200$) renders a smaller
relative difference in PMSE, i.e. $\textrm{PMSE}_{\textrm{EBcoBART}}/\textrm{PMSE}_{\textrm{BART}}$
increases, compared to a smaller sample sizes ($N=100$). This finding
is expected because prior information is more relevant for smaller
sample sizes. 

A comparison between the two tree flexibility settings reveals that
the flexible tree models benefit more from the co-data than the rigid
tree models (\autoref{fig:RatioPMSE}). The flexible model uses more nodes and can therefore relate better to the co-data information. Moreover, the co-data helps to prune away non-informative covariates, which are less frequent for
the rigid model. 

\subsection{Dense and linear setting}
\label{subsec:Dense-and-linear}

We briefly discuss results for a dense, linear regression simulation
setting with continuous co-data. Details are found in the Supplement
(Section 5). For variable importance, results are similar as for the sparse nonlinear simulation: EB-coBART upweights predictive covariates and downweights nonpredictive covariates. This up/down-weighting effect is stronger for rigid BART model as this model
favors variable selection compared to the flexible model (Figure S5). 

For most data sets, EB-coBART has a lower PMSE than BART.
This effect is stronger for the flexible models. The flexible model has a lower average PMSE compared to the rigid model for all simulation settings, as expected. For dense covariate-response relationships, many splitting rules are required, which are more easily captured by a flexible model compared to a rigid model (Figure S6).

The linear models Ecpc and ridge regression outperform all BART models because the data generating mechanism is linear. Ecpc benefits from the co-data, as evidenced by a substantial decrease in PMSE compared to ridge regression. Among the nonlinear models, flexible EB-coBART has the lowest average PMSE, and specifically a lower PMSE than random forest and CoRF (Table S1). 

\subsection{Uninformative co-data}
\label{subsec:Uninformative-co-data}

Details for a setting with uninformative grouping co-data are found
in Supplementary Section 6 (Figures S7 and S8). Briefly, EB-coBART and BART perform similar in this co-data setting. In the rigid tree setting, EB-coBART does not upweight the groups on average, but there are fluctuations in the estimated group weights. These fluctuations also induce fluctuations in the test performance ratio, although on average both BART models have an almost equal PMSE.

In the flexible tree setting, EB-coBART and BART are practically
identical. EB-coBART does not upweight the groups on average and fluctuations
in the estimated group weights across the data sets are hardly present.
The PMSE of EB-coBART and BART are close to equal for all data sets.

\section{Application}
\label{sec:Applications}

\paragraph*{Data} 
We apply our method to the prognosis of Diffuse large B-cell lymphoma
(DLBCL) patients. DLBCL is a common non-Hodgkin lymphoma for which accurate prognosis is difficult because of the clinical and biological heterogeneity of the patients. One well-accepted prognostic clinical covariate is the international prognostic index (IPI), which scores patients based on their age, the stage of the tumor, lactose
dehydrogenase levels, a mobility measure, and the number of extranodal
sites (\cite{IPI}). However, the predictive power of IPI is still limited.
Therefore, a major branch of DLBCL research focuses on finding new
omics-based markers for DLBCL prognosis.

To this end, we fit EB-coBART to a cohort of $101$ uniformly treated DLBCL
patients for whom we aim to predict two year progression-free survival
(yes/no, $18\%$ no). We treat the outcome as binary because two year is a clinically well-accepted cut-off and censoring was absent within this time period. We also have, for each patient, complete measurements
of a total of $p=140$ covariates divided in four types: $67$ DNA copy
number variations (CNV), $69$ point mutations of genes, $3$ translocations,
and $1$ clinical covariate: IPI.

We provide EB-coBART with the following two co-data sources. First,
because we have different types of covariates which may have different
scales in relation to the response, we group covariates by type (CNV, mutation, translocation, or clinical). IPI is
a group on its own because of its known prognostic importance. Second,
we provide continuous co-data by estimating Benjamini-Hochberg corrected
p-values ($-\textrm{logit}$ scale) of all covariates in association
with the binary response from a large ($N=430$) previously published
cohort with DLBCL patients having received a slightly different treatment
compared to the patients in the training cohort. In this way, EB-coBART makes effective use of available data, while still acknowledging that the effect of prognostic factors may depend on the type of treatment. 

\paragraph*{Analysis} 
We fit EB-coBART models to the training cohort. We consider EB-coBART initialized in the flexible tree setting ($\alpha=0.95,$ $\beta=2,$ $k=2$),  which are the default settings proposed by \citet{chipmanBART2010}, and EB-coBART initialized in the rigid tree setting ($\alpha=0.95,$ $\beta=2,$ $k=2$). Additionally, we consider, for both tree settings, two hyperparameter estimation strategies. The first strategy (EB-coBART 1) only estimates the covariate weight hyperparameter $\boldsymbol{S}$ of BART by estimator \eqref{eq:co-data EB estimates of S}, similar to \autoref{sec:Simulations}, and the second strategy (EB-coBART 2) simultaneously estimates $\boldsymbol{S}$ by \eqref{eq:co-data EB estimates of S} and hyperparameters ($\alpha,$ $k$) according to \eqref{eq:HyperparameterUpdatesTree} and \eqref{eq:HyperparameterUpdatesLeaf}, with ($\alpha,$ $k$) initialized in the flexible or the rigid tree setting. Thus, we have a total of four EB-coBART models: flexible EB-coBART 1, flexible EB-coBART 2, rigid EB-coBART 1, and rigid EB-coBART 2. Hyperparameter $\beta$ is fixed to $\beta=2$ for flexible models and $\beta=4$ for the rigid models.

Because BART uses a probit link for binary responses (\cite{albert1993bayesian}), we have $p(y_{i}=1\mid\boldsymbol{x}_{i})=\Phi\left(f(\boldsymbol{x}_{i})\right),$
with $\Phi$ the standard normal cdf and $f\left(\boldsymbol{x}_{i}\right)$
the latent response, which will be modeled by a sum-of-trees. Therefore, the
error variance equals $\sigma^{2}=1$ and  hyperparameters
$\left(\nu,\lambda\right)$ for $\pi\left(\sigma^{2}\right)$ are
not required. Additionally, probit BART has terminal node value prior $\mu_{lt}\sim\mathcal{N}\left(0,3/(k\sqrt{K})\right)$ instead of \eqref{eq:LeafNode prior}. We fit BART models using $10$ chains each consisting
of $24000$ samples of which $12000$ are burn-in.

We evaluate predictive performance by the area under the curve (AUC), estimated using the R package pROC (\cite{pROC}), and the average Brier score: $\textrm{Brier}= N^{-1}\sum_{i=1}^{N}\left(\hat{y}_{i}-y_{i}\right)^{2}$. First, we estimate the performance internally on the training cohort using repeated ($3\times$) $10$-fold cross-validation (CV). Second, we fit the models on the full training cohort and  estimate the performance on an external test cohort consisting of $83$ patients with the same treatment. In addition, we investigate the predictive performance of EB-coBART 2 as a function of the sample size by estimating a learning curve (\cite{LearningCurves}).  

We compare EB-coBART with flexible and rigid BART, i.e. BART models having equal covariate weights ($s_{j}=1/p$), and with cv-BART, which estimates hyperparameters $\alpha,$ $k,$ and the number of trees $K$ using $5$-fold CV, and fixes $\beta=2$ and $s_{j}=1/p.$ We consider the CV
grid $\alpha=\{0.1,\,0.5,\,0.95\},$ $k=\{1,\,2,\,3\}$,
and $K=\{50,\,150\}.$ We also fit BART using only IPI as covariate (IPI-BART). 

Next to the BART models, we include a comparison with random forest
and its co-data extension CoRF (both fitted using $2000$ trees with
the R package randomForestSRC (\cite{randomforestSRC}), and to ridge
regression and its co-data extension Ecpc (\cite{ecpc,ecpcpackage}).
For Ecpc, we also considered a posthoc variable selection with the
optimal number of covariates $p_{\textrm{sel}}=\{2,5,10,50,80\}$,
but this did not render improved performances. 

\paragraph*{Results}
Predictive performance results are shown in \autoref{tab:PredictivePerfAplication}.
To facilitate comparison, and because the flexible tree setting ($\alpha=0.95,$ $\beta=2,$ $k=2$) is the default (\cite{chipmanBART2010}), we show results for this setting here. Results for the rigid tree setting ($\alpha=0.1,$ $\beta=4,$ $k=1$) (Supplementary Table S2) are summarized.

\begin{table}[h!]
\centering{}\caption{Predictive performance estimates, based on repeated ($3\times$) $10$-fold cross-validation and an external test cohort, of several prediction models.}
\label{tab:PredictivePerfAplication}
\begin{tabular}{ccccc}
 & \multicolumn{2}{c}{\textbf{Cross-Validation}} & \multicolumn{2}{c}{\textbf{Test Set Cohort}}\tabularnewline
 & AUC & Brier score & AUC & Brier score\tabularnewline
\hline 
\hline 
flexible BART & 0.68 & 0.173 & 0.557 & 0.162\tabularnewline
cv-BART & 0.68 & 0.168 & 0.557 & 0.162\tabularnewline
flexible EB-coBART 1 & 0.72 & 0.155 & 0.697 & 0.154\tabularnewline
flexible EB-coBART 2 & 0.71 & 0.156 & \textbf{0.714} & \textbf{0.153}\tabularnewline
IPI-BART & 0.70 & \textbf{0.153} & 0.669 & 0.154\tabularnewline
Random Forest & 0.69 & 0.183 & 0.63 & 0.158\tabularnewline
CoRF & 0.71 & 0.157 & 0.65 & 0.159\tabularnewline
Ecpc & \textbf{0.73} & 0.159 & 0.705 & 0.154\tabularnewline
Ridge & 0.72 & 0.161 & 0.69 & 0.154\tabularnewline
\end{tabular}
\end{table}

For the cross-validated performances, the prediction models are competitive. EB-coBART has a marginally improved performance compared to BART and cv-BART in both tree flexibility settings, while IPI-BART performs similarly to EB-coBART.

Compared to the cross-validated performance, the differences between the models are larger when evaluated on the test set. EB-coBART 2 has
the best performance, and slightly better than EB-coBART 1. Because EB-coBART 2 estimates $\alpha=0.61$, it renders a sparser tree model compared to EB-coBART 1, which uses $\alpha=0.95.$ The differences with BART and cv-BART, which yields $k=2,$
$\alpha=0.5,$ and $K=150$ as hyperparameter estimates, is large.
These differences are significant using DeLong's paired
test (\cite{delong1988comparing}) for the difference in AUC ($p_{\varDelta AUC}=0.013$, $p_{\varDelta AUC}=0.024,$ respectively) and the Wilcoxon signed
rank test for the difference in brier score ($p_{\varDelta Brier}=0.034$,
$p_{\varDelta Brier}=0.0074,$ respectively). IPI-BART
also substantially outperforms BART and cv-BART.

EB-coBART has a substantially, but not significantly, better test performance
than random forest and CoRF, while Ecpc and ridge regression are competitive.
Ecpc and CoRF benefit only marginally from the co-data, indicated by the comparative performance of their corresponding base learners. 

Rigid EB-coBART has a comparable performance to flexible EB-coBART (Supplementary Table S2), whereas rigid BART (without co-data) performs better than the standard, i.e. flexible BART. This suggests a sparse covariate-response relationship for this application, which, unlike for the standard BART, is picked up by flexible EB-coBART as well due to the informative co-data.

\begin{figure}[h!]
\begin{centering}
\subfloat[\label{fig:LearningCurves}]{\includegraphics[scale=0.7]{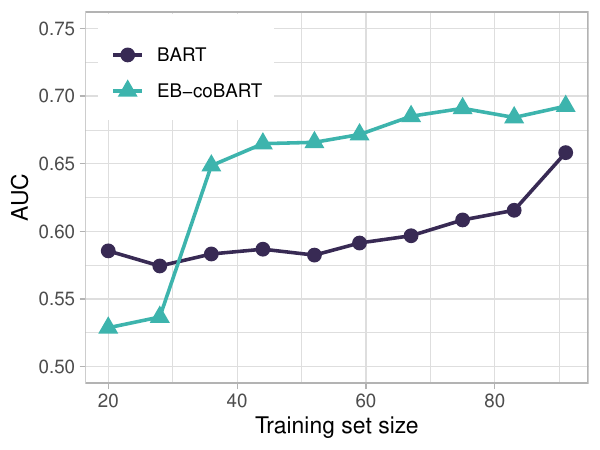}
}\subfloat[\label{fig:WAIC}]{\includegraphics[scale=0.7]{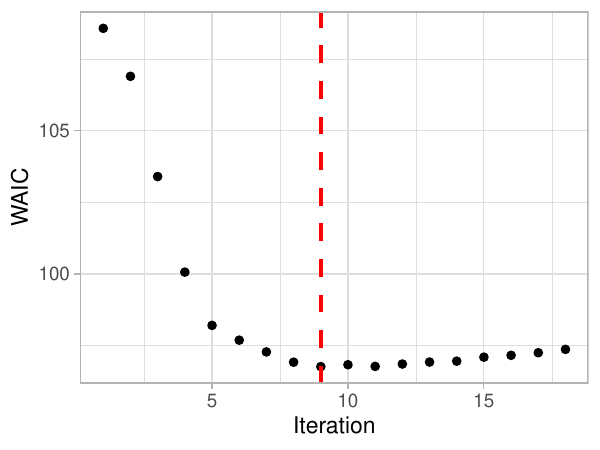}
}
\par\end{centering}
\begin{centering}
\subfloat[\label{fig:GroupWeights}]{\includegraphics[scale=0.7]{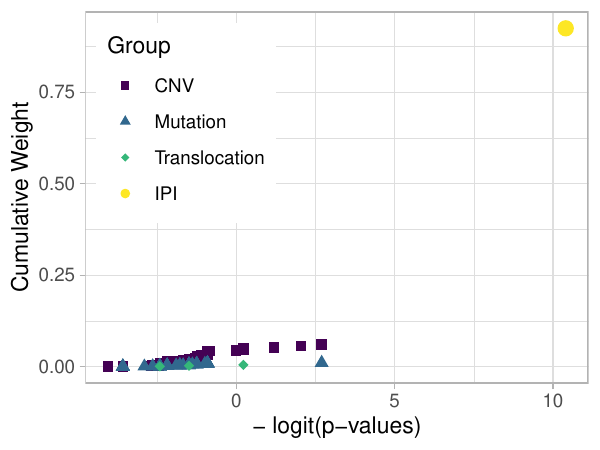}}\subfloat[\label{fig:PartialDependence}]{\includegraphics[scale=0.7]{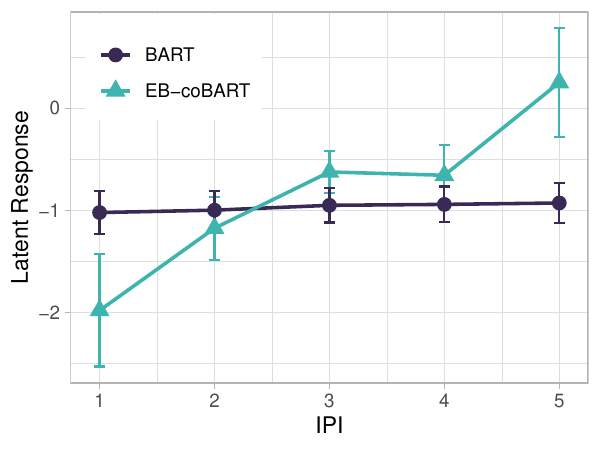}
}\caption{\small{\textbf{(a)} Learning curves for EB-coBART 2 (triangles)
and BART (dots). \textbf{(b)} Estimated WAIC (dots) for $18$ iterations with the minimum indicated by the dashed vertical line at iteration $9.$ \textbf{(c)} Estimated cumulative covariate weights for the four types of covariates as a function of external p-values on the $-\textrm{logit}$ scale for EB-coBART 2. Types of covariates are indicated by a square (copy number variation), a triangle (mutation), a diamond (translocation), and a circle (IPI). \textbf{(d)} Partial dependence plots of EB-coBART 2 (triangles) and BART (dots) showing the marginal effect of IPI on the predictions. On the y-axis, we show the latent response values, i.e. $Z$-values of the standard normal cdf, because BART models use a probit link for binary responses. We show the average $\pm$ the standard deviation across the Gibbs samples of the latent response.}}
\label{fig:ApplicationResults}
\par\end{centering}
\end{figure}

The learning curves for flexible EB-coBART 2 (triangles) and flexible BART (dots)
are depicted in \autoref{fig:LearningCurves}. Learning curves are created as follows.
For ten different subsample sizes $n$, we sample multiple training sets without replacement from the training cohort and define corresponding test sets as left-out samples. We then fit flexible EB-coBART 2 (updating $s_{j},$ $\alpha,$ $k$) and flexible BART to the training sets and estimate the AUC on the complementary test sets. We aggregate AUC estimates per subsample size by the average and plot them to obtain the learning curve. 

For small subsample size $n$, BART has a larger AUC than EB-coBART. Moving from $n=28$ to $n=36,$ the AUC of EB-coBART shows a phase transition
(\cite{PhaseTransition}) by jumping to a much larger AUC value. This
transition is also present in the estimated weight for IPI, which makes a large jump at this transition subsample size (Supplementary Figure S9). The AUC of BART
remains relatively constant across the sample size trajectory and only
shows a clear increase in AUC at the end. This indicates that the inclusion of co-data in the BART algorithm is most beneficial for medium sample sizes, as the sample size is then large enough to pick up the co-data signal,  while still too small to be picked up without the co-data guidance. 

EB-based estimates of hyperparameter $\alpha$ increase for larger
sample sizes (Supplementary Figure S10a),
because less tree structure shrinkage is required when more information
is available. Hyperparameter $k$ stayes relatively stable across
different subsample sizes (Supplementary Figure S10b).

Because flexible EB-coBART 2 (updating $\boldsymbol{S},$ $\alpha,$ $k$) performs best on the test set, we show this fit in \autoref{fig:ApplicationResults}b and c, and discuss differences with EB-coBART 1 and rigid EB-coBART. \autoref{fig:WAIC} shows estimates of the WAIC across $18$ iterations, and \autoref{fig:GroupWeights}
shows co-data moderated EB estimates of the covariate weights $\hat{w}_{j}^{(q)}$, \eqref{eq:co-data EB estimates of S}, aggregated by covariate type, at minimum WAIC. 

The estimated WAIC of EB-coBART 2 has a clear minimum at iteration
$9$ (\autoref{fig:WAIC}), which is similar to EB-coBART 1 and the rigid models. At minimum WAIC, EB-coBART 2 shows a strong preference for IPI, which receives $92.5$\% of the total weight (\autoref{fig:GroupWeights}). The remaining weight is spread out thinly among the omics covariates, with the sum of the weights per type equal to $6.00$\% for the copy number variations, $1.01$\% for the mutations, and $0.49$\% for the translocations. IPI receives about $97$\% of the weight when rigid EB-coBART is considered. Devoting much weight to IPI indicates that EB-coBART automatically finds a relevant signal, because BART fitted with just IPI (IPI-BART) has a substantially better performance than BART (\autoref{tab:PredictivePerfAplication}).

Because IPI appears to be dominant for this application, we show
the marginal effect of IPI on the predictions of EB-coBART 2 and BART by the partial dependence function (\cite{PartialDependence}) (\autoref{fig:PartialDependence}). Marginalized IPI shows a clear effect on the two year progression
free survival (PFS) predictions of EB-coBART, where larger IPI values decrease
the probability of two year PFS. Furthermore, the effect is slightly
nonlinear. The signal in IPI is absent for BART.

\section{Discussion}
\label{sec:Discussion}

We developed EB-coBART, a method that incorporates co-data into BART by estimating prior covariate weights using empirical Bayes and a co-data model. This method rendered improved predictions, depending on the informativeness of the co-data, compared to standard BART in simulations and in an application to lymphoma patients. Furthermore, this application illustrated that EB-coBART performs better than CoRF (\cite{teBeest2017codata}) and competitively to Ecpc (\cite{ecpc}), two state-of-the-art co-data learners. The comparative performance of EB-coBART and regression-based Ecpc depends on how well the true model can be approximated by a linear one.

We considered EB-coBART having a flexible tree setting, which is the default proposed by \citet{chipmanBART2010}, and a rigid tree setting. The rigid model is sparser because it uses less nodes and it therefore lends itself better for variable selection and model interpretability compared to the flexible model. For prediction, we therefore recommend the rigid model when the performance is comparable, as was the case in the application.

The application illustrated that the specification of co-data may be rather straightforward, as a simple grouping by covariate type already rendered informative co-data for BART. Furthermore, the combination of empirical Bayes and a parsimonious co-data model protects against overfitting when uninformative co-data is incorporated. This parsimony may be slightly relaxed by considering nonlinear co-data models such as general additive models (\cite{GAMS}). Success of such strategy depends on the number of covariates, as these are effectively the number of samples in the co-data model. Furthermore, additional penalization in this model may be required to avoid overfitting (\cite{ecpc}).

EB-coBART has to refit BART at each iteration and it is therefore computationally intensive. However, because we empirically found that $4-12$ iterations is typical, EB-coBART requires less computational time than cross-validated BART. The computational time of BART, and thus also of EB-coBART, does not scale well to very high-dimensional settings ($p\approx10^{4}$). It may therefore be interesting to apply our method to BART-BMA (\cite{BARTBMA}), a computationally efficient BART-based method. However, BART-BMA employs greedy search instead of having a prior on the covariate weights, which makes the implementation of empirical Bayes not straightforward.

We end with suggestions for future research. First, some co-data presents itself in a hierarchical structure, such as gene ontology trees. Such structures may possibly be accommodated by EB-coBART in the co-data model, e.g. by borrowing ideas from the hierarchical lasso (\cite{HierarchicalLasso}). Second, for tree-based methods, it is generally difficult to deal with types of covariates that have a different scale and priority in relation to the response. Grouping the covariates by type provided a simple solution, but other, more sophisticated, solutions may improve the tree-based model. It may be good to favor certain types, e.g. clinical covariates, at the trunk of the trees, as such covariates may be easier to interpret and may have proven their use more extensively. Thirdly, we currently did not account for the depth at which covariates occur when fitting the co-data model; we only accounted for their occurrence. While the latter may suffice for settings with fairly large $p$, it may be too granular for low $p$ settings. Depths may be used as inverse weights in the co-data model to give more weight to covariates close to the tree's trunk.

\section{Data availability and software}
\label{Sofware}
For the DLBCL application, treatment and covariate names were anonymized, as the original data has not been published yet. Anonymized data includes the training cohort, the test cohort, and the co-data matrix. These data and R code (version version 4.3.0) to reproduce results presented in \autoref{sec:Simulations}  and \autoref{sec:Applications} are available via \url{https://github.com/JeroenGoedhart/EB_coBART_paper}.

\section{Supplementary Material}
\label{Supplements}
Supplementary material is available

\section*{Acknowledgments}
The authors thank Bauke Ylstra and Daphne de Jong for providing and discussing the data presented in \autoref{sec:Applications}.
This research was funded by Hanarth Fonds.

\noindent{\it Conflict of Interest}: None declared.

\begin{singlespace}

\end{singlespace}
\end{document}


\sloppy
\begin{center}
\large\textbf{SUPPLEMENTARY MATERIAL TO: Adaptive Use of Co-data through Empirical Bayes for Bayesian Additive Regression Trees}
\end{center}\bigskip{}

\begin{center}
Jeroen M. Goedhart$^\ast$\textsuperscript{1}, Thomas Klausch\textsuperscript{1}, Jurriaan Janssen\textsuperscript{2}, Mark A. van de Wiel\textsuperscript{1}
\end{center}

\begin{center}
$^\ast$\small{Correspondence e-mail address: j.m.goedhart@amsterdamumc.nl}
\end{center}

\noindent\textsuperscript{1}\footnotesize{Department of Epidemiology \& Data Science, Amsterdam Public Health Research Institute,
Amsterdam University Medical Centers Location AMC, Meibergdreef 9,
the Netherlands}\\
\\
\textsuperscript{2}\footnotesize{Department of Pathology, Cancer Center Amsterdam,
Amsterdam University Medical Centers Location VUMC,
De Boelelaan 1117, the Netherlands}

\section{Sampling from the posterior of BART}
\label{sec:Sampling-from-the}
The posterior of BART is proportional to
\small
\begin{align}
\pi\left(\boldsymbol{\mathcal{T}},\boldsymbol{\mathcal{M}},\sigma^{2}\mid\boldsymbol{y},\boldsymbol{X};\alpha,\beta,k,\nu,\lambda,\boldsymbol{S}\right) & \propto\left[\prod_{i=1}^{N}\mathcal{N}\left(Y_{i};G\left(\boldsymbol{x}_{i};\boldsymbol{\mathcal{T}},\boldsymbol{\mathcal{M}}\right),\sigma^{2}\right)\right]\label{eq:full posterior-1}\\
 & \times\left[\prod_{t=1}^{K}\prod_{l=1}^{L_{t}}\pi\left(\mathcal{T}_{t};\alpha,\beta,\boldsymbol{S}\right)\mathcal{N}\left(\mu_{lt};0,\frac{0.5}{k\sqrt{K}}\right)\right]\nonumber\\
 & \times\left[\mathcal{IG}\left(\sigma^{2};\frac{\nu}{2},\frac{\nu\lambda}{2}\right)\right].\nonumber 
\end{align}
\normalsize
To sample from the posterior, \eqref{eq:full posterior-1} is first
decomposed into two full posterior conditionals:
\small
\begin{align}
\pi\left(\boldsymbol{\mathcal{T}},\boldsymbol{\mathcal{M}}\mid\boldsymbol{y},\boldsymbol{X},\sigma^{2};\alpha,\beta,k,\boldsymbol{S}\right) & =\pi\left(\boldsymbol{\mathcal{T}},\boldsymbol{\mathcal{M}}\mid\boldsymbol{y},\boldsymbol{X};\alpha,\beta,k,\boldsymbol{S}\right)\label{eq:Conditionals of BART}\\
\pi\left(\sigma^{2}\mid\boldsymbol{\mathcal{T}},\boldsymbol{\mathcal{M}},\boldsymbol{y},\boldsymbol{X};\nu,\lambda\right) & =\mathcal{IG}\left(\sigma^{2};\frac{N+\nu}{2},\frac{\nu\lambda+\sum_{i=1}^{N}\left(y_{i}-\hat{y}_{i}\right)^{2}}{2}\right)\nonumber 
\end{align}
\normalsize
The last line of \eqref{eq:Conditionals of BART} originates from
the conjugacy of the inverse gamma prior. The first line of \eqref{eq:Conditionals of BART} requires further decomposition. 

This decomposition is performed using a Bayesian backfitting algorithm
proposed by \cite{hastie2000bayesianbackfitting}. By noting that a tree $t$
depends on all other trees only through the residual response $\boldsymbol{r}_{t}=\boldsymbol{y}-\sum_{t'\neq t}^{K-1}g\left(\boldsymbol{X};\mathcal{T}_{t},\mathcal{M}_{t'}\right),$
we may simplify a draw from the first line of \eqref{eq:Conditionals of BART}
to $K$ draws of subsequent trees. A single tree Monte Carlo sample
consists of first sampling a new tree structure $\mathcal{T}_{t}$
and then an update of the terminal node parameters $\mathcal{M}_{t}:$
\small
\begin{align}
\pi\left(\mathcal{T}_{t}\mid\boldsymbol{r}_{t},\sigma^{2};\alpha,\beta\right) & \propto\pi\left(\mathcal{T}_{t};\alpha,\beta\right)p\left(\boldsymbol{r}_{t}\mid\mathcal{T}_{t},\sigma^{2}\right)\label{eq:Tree sampling Scheme}\\
\pi\left(\mathcal{M}_{t}\mid\mathcal{T}_{t},\boldsymbol{r}_{t},\sigma^{2};k\right) & =\prod_{l=1}^{L_{t}}\pi\left(\mu_{lt}\mid\mathcal{T}_{t},\boldsymbol{r}_{lt},\sigma^{2};k\right).\nonumber 
\end{align}
\normalsize
A Metropolis-Hasting algorithm is employed to sample new tree structures
$\mathcal{T}_{t}$ (first line of \eqref{eq:Tree sampling Scheme}).
Here, a local tree modification proposal distribution is used for
which full details are described by \cite{tan2019bayesiantutorial,KapelnerBART}.
The conjugacy of the terminal node prior (eq. 2.6 of main text)
ensures that the second line of \eqref{eq:Tree sampling Scheme}
is a $\mathcal{N}\left(\mu_{lt};\frac{\sigma^{-2}\sum_{h=1}^{n_{lt}}r_{hlt}}{n_{lt}/\left(\sigma^{2}+\sigma_{\mu}^{2}\right)},\frac{1}{n_{lt}/\left(\sigma^{2}+\sigma_{\mu}^{2}\right)}\right),$
with $n_{lt}$ the number of observations falling in the $l$th terminal
node of tree $t,$ and $h$ an index variable representing the $h$th
observation of a given terminal node and tree. 

\section{Stochastic EM algorithm}
\label{sec:Stochastic-EM-algorithm}

Here, we derive the stochastic EM algorithm for marginal likelihood
maximization w.r.t. the hyperparameters for BART. Let $\boldsymbol{\psi}$ be a vector of hyperparameters and $\textrm{ML}\left(\boldsymbol{y}\mid\boldsymbol{X};\boldsymbol{\psi}\right)$
the marginal likelihood as a function of $\boldsymbol{\psi}.$ In
Empirical Bayes, the goal is to find the hyperparameters that satisfy
\small
\begin{equation}
\hat{\boldsymbol{\psi}}=\argmax_{\boldsymbol{\psi}}\;\textrm{ML}\left(\boldsymbol{y}\mid\boldsymbol{X};\boldsymbol{\psi}\right).\label{eq:marglik optimization}
\end{equation}
\normalsize
To approximate the optimization in \eqref{eq:marglik optimization},
\cite{Casella2001empirical} proposed a stochastic EM algorithm 
that iteratively uses Gibbs samples from the posterior $\pi_{\boldsymbol{\psi}}\left(\boldsymbol{\theta}\mid\boldsymbol{y},\boldsymbol{X}\right)$.
This algorithm reads in general form
\small
\[
\boldsymbol{\hat{\psi}}^{(q+1)}=\argmax_{\boldsymbol{\psi}}\;E_{\pi\left(\boldsymbol{\theta}\mid\boldsymbol{y},\boldsymbol{X};\boldsymbol{\hat{\psi}}^{(q)}\right)}\left[\log \ell\left(\boldsymbol{\theta},\boldsymbol{y}\mid\boldsymbol{X};\boldsymbol{\psi}\right)\right]\approx\argmax_{\boldsymbol{\psi}}\;\frac{1}{n_{mc}}\sum_{m=1}^{n_{mc}}\log \ell\left(\boldsymbol{\theta}_{m}^{(q)},\boldsymbol{y}\mid\boldsymbol{X};\boldsymbol{\psi}\right),
\]
\normalsize
with $q$ the iteration index and $\ell\left(\boldsymbol{\theta},\boldsymbol{y}\mid\boldsymbol{X};\boldsymbol{\psi}\right)$ the conditional likelihood of $\boldsymbol{\psi}$, i.e. it evaluates the joint probability of data $\boldsymbol{y}\mid\boldsymbol{X}$ and model parameters $\boldsymbol{\theta}$ as a function of $\boldsymbol{\psi}.$ The expectation of this conditional likelihood is taken w.r.t to the posterior distribution with hyperparameter $\boldsymbol{\psi}^{(q)}.$
Then, for the right-hand side, we have $m$ the $m$th Monte Carlo sample, $n_{mc}$ the total number
of Monte Carlo samples, $q$ the $q$th iteration, and $\ell\left(\boldsymbol{\theta}_{m}^{(q)},\boldsymbol{y}\mid\boldsymbol{X};\boldsymbol{\psi}\right)$
the conditional likelihood of $\boldsymbol{\psi}$ evaluated at the $m$th posterior sample of the model parameters $\boldsymbol{\theta}_{m}^{(q)}.$
Specifically, $\boldsymbol{\theta}_{m}^{(q)}$
denotes the $m$th model parameter sample from the posterior $\pi\left(\boldsymbol{\theta}\mid\boldsymbol{y},\boldsymbol{X};\boldsymbol{\hat{\psi}}^{(q)}\right)$
with the hyperparameters set to $\boldsymbol{\hat{\psi}}^{(q)}.$

For BART, the model parameters are $\boldsymbol{\theta}=\left(\boldsymbol{\mathcal{T}},\boldsymbol{\mathcal{M}},\sigma\right),$
and the hyperparameters are $\boldsymbol{\psi}=\left(\alpha,\beta,k,\nu,\lambda,\boldsymbol{S}\right),$
which leads to
\small 
\begin{equation}
\boldsymbol{\hat{\psi}}^{(q+1)}\approx\arg\max_{\boldsymbol{\psi}}\frac{1}{n_{mc}}\sum_{m=1}^{n_{mc}}\log \ell\left(\boldsymbol{y},\boldsymbol{\mathcal{T}}_{m}^{(q)},\boldsymbol{\mathcal{M}}_{m}^{(q)},\sigma_{m}^{2(q)}\mid\boldsymbol{X};\boldsymbol{\psi}\right),\label{eq:EBupdate BART}
\end{equation}
\normalsize
with $\left(\boldsymbol{\mathcal{T}}_{m}^{(q)},\boldsymbol{\mathcal{M}}_{m}^{(q)},\sigma_{m}^{2(q)}\right)$
denoting the $m$th posterior sample of the tree parameters and the
error variance at iteration $q$. By factoring out the joint likelihood
and dropping all terms that do not depend on the hyperparameters,
\eqref{eq:EBupdate BART} can be rewritten as
\small
\begin{align}
\boldsymbol{\hat{\psi}}^{(q+1)} & \approx\underset{\alpha,\beta,k,\nu,\lambda,\boldsymbol{S}}{\arg\max}\frac{1}{n_{mc}}\sum_{m=1}^{n_{mc}}\log\left[\pi\left(\boldsymbol{y}\mid\boldsymbol{X},\boldsymbol{\mathcal{T}}_{m}^{(q)},\boldsymbol{\mathcal{M}}_{m}^{(q)},\sigma_{m}^{2(q)}\right)\pi\left(\boldsymbol{\mathcal{T}}_{m}^{(q)},\boldsymbol{\mathcal{M}}_{m}^{(q)},\sigma_{m}^{2(q)};\alpha,\beta,k,\nu,\lambda,\boldsymbol{S}\right)\right]\label{eq:em for marginal likelihood bart}\\
 & =\underset{\alpha,\beta,k,\nu,\lambda,\boldsymbol{S}}{\arg\max}\sum_{m=1}^{n_{mc}}\log\left[\pi\left(\boldsymbol{\mathcal{T}}_{m}^{(q)},\boldsymbol{\mathcal{M}}_{m}^{(q)},\sigma_{m}^{2(q)};\alpha,\beta,k,\nu,\lambda,\boldsymbol{S}\right)\right],\nonumber 
\end{align}
\normalsize
with $\pi\left(\boldsymbol{\mathcal{T}}_{m}^{(q)},\boldsymbol{\mathcal{M}}_{m}^{(q)},\sigma_{m}^{2(q)};\alpha,\beta,k,\nu,\lambda,\boldsymbol{S}\right)$
the prior probabilities of the $m$th posterior sample $\left(\boldsymbol{\mathcal{T}}_{m}^{(q)},\boldsymbol{\mathcal{M}}_{m}^{(q)},\sigma_{m}^{2(q)}\right)$,
with hyperparameters $\boldsymbol{\hat{\psi}}^{(q)}.$ Equation \ref{eq:em for marginal likelihood bart}
is further decomposed as
\small
\[
\boldsymbol{\hat{\psi}}^{(q+1)}=\underset{\alpha,\beta,k,\nu,\lambda,\boldsymbol{S}}{\arg\max}\sum_{m=1}^{n_{mc}}\left[\log\pi\left(\boldsymbol{\mathcal{T}}_{m}^{(q)};\alpha,\beta,\boldsymbol{S}\right)+\log\pi\left(\boldsymbol{\mathcal{M}}_{m}^{(q)}\mid\boldsymbol{\mathcal{T}}_{m}^{(q)};k\right)+\log\pi\left(\sigma_{m}^{2(q)}\mid\boldsymbol{\mathcal{T}}_{m}^{(q)},\boldsymbol{\mathcal{M}}_{m}^{(q)};\nu,\lambda\right)\right],
\]
\normalsize
 thus rendering
\small
\begin{align}
\left(\hat{\alpha}^{(q+1)},\hat{\beta}^{(q+1)},\boldsymbol{\hat{S}}^{(q+1)}\right) & =\underset{\alpha,\beta,\boldsymbol{S}}{\arg\max}\sum_{m=1}^{n_{mc}}\log\pi\left(\boldsymbol{\mathcal{T}}_{m}^{(q)};\alpha,\beta,\boldsymbol{S}\right)\label{eq:EBupdate tree structure}\\
\hat{k}^{(q+1)} & =\underset{k}{\arg\max}\sum_{m=1}^{n_{mc}}\log\pi\left(\boldsymbol{\mathcal{M}}_{m}^{(q)}\mid\boldsymbol{\mathcal{T}}_{m}^{(q)};k\right)\label{eq:EBupdate leaf node}\\
 & =\arg\max_{k}\sum_{m=1}^{n_{mc}}\log\left(\prod_{t=1}^{K}\prod_{l=1}^{L_{tm}^{(q)}}\mathcal{N}\left(\mu_{ltm}^{(q)};0,\sigma_{\mu}^{2}\right)\right)\nonumber \\
 & =\underset{k}{\arg\max}\sum_{m=1}^{n_{mc}}\sum_{t=1}^{K}\sum_{l=1}^{L_{tm}^{(q)}}\log\mathcal{N}\left(\mu_{ltm}^{(q)};0,\frac{3}{k\sqrt{K}}\right)\nonumber \\
\left(\hat{\nu}^{(q+1)},\hat{\lambda}^{(q+1)}\right) & =\underset{\nu,\lambda}{\arg\max}\sum_{m=1}^{n_{mc}}\log\mathcal{IG}\left(\sigma_{m}^{2(q)};\frac{\nu}{2},\frac{\nu\lambda}{2}\right),\label{eq:EBupdate error variance}
\end{align}
\normalsize
with $L_{tm}^{(q)}$ the number of terminal nodes of tree $t$ in
Monte Carlo sample $m$ at iteration $q,$ and $\mu_{ltm}^{(q)}$
the $l$th sampled terminal node of tree $t$ in the $m$th Monte
Carlo sample at iteration $q.$ Now, \eqref{eq:EBupdate leaf node}
and \eqref{eq:EBupdate error variance} are standard maximum likelihood
estimation problems for the normal and inverse gamma distribution,
respectively. Equation \ref{eq:EBupdate tree structure} requires
\footnotesize
\begin{align*}
\left(\hat{\alpha}^{(q+1)},\hat{\beta}^{(q+1)},\boldsymbol{\hat{S}}^{(q+1)}\right) & =\underset{\alpha,\beta,\boldsymbol{S}}{\arg\max}\sum_{m=1}^{n_{mc}}\log\prod_{t=1}^{K}\left[\prod_{z=1}^{Z_{tm}^{(q)}}\textrm{Categorical}\left(j_{ztm}^{(q)};\boldsymbol{S}\right)\prod_{z=1}^{Z_{tm}^{(q)}}\alpha\left(1+d_{ztm}\right)^{-\beta}\prod_{l=1}^{L_{tm}^{(q)}}1-\alpha\left(1+d_{ltm}\right)^{-\beta}\right],\\
 & =\underset{\boldsymbol{S}}{\arg\max}\sum_{m=1}^{n_{mc}}\sum_{t=1}^{K}\log\left[\prod_{z=1}^{Z_{tm}^{(q)}}\textrm{Categorical}\left(j_{ztm}^{(q)};\boldsymbol{S}\right)\right]\\
 & \quad+\underset{\alpha,\beta}{\arg\max}\sum_{m=1}^{n_{mc}}\sum_{t=1}^{K}\log\left[\prod_{z=1}^{Z_{tm}^{(q)}}\alpha\left(1+d_{ztm}^{(q)}\right)^{-\beta}\prod_{l=1}^{L_{tm}^{(q)}}1-\alpha\left(1+d_{ltm}^{(q)}\right)^{-\beta}\right]
\end{align*}
\normalsize
which renders the updates 
\small
\begin{align}
\boldsymbol{\hat{S}}^{(q+1)} & =\underset{\boldsymbol{S}}{\arg\max}\sum_{m=1}^{n_{mc}}\sum_{t=1}^{K}\sum_{z=1}^{Z_{tm}^{(q)}}\log\left[\textrm{Categorical}\left(j_{ztm}^{(q)};\boldsymbol{S}\right)\right]\label{eq:EBupdate split variable}\\
\left(\hat{\alpha}^{(q+1)},\hat{\beta}^{(q+1)}\right) & =\underset{\alpha,\beta}{\arg\max}\sum_{m=1}^{n_{mc}}\sum_{t=1}^{K}\log\left[\prod_{z=1}^{Z_{tm}^{(q)}}\alpha\left(1+d_{ztm}^{(q)}\right)^{-\beta}\prod_{l=1}^{L_{tm}^{(q)}}1-\alpha\left(1+d_{ltm}^{(q)}\right)^{-\beta}\right],\label{eq:EBupdate alpha beta}
\end{align}
\normalsize
with $j_{ztm}$ denoting splitting variable $x_{j}$ of the $z$th
internal node of tree $t$ and Monte Carlo sample $m.$ Equation \ref{eq:EBupdate split variable}
is recognized as maximum likelihood estimation of a categorical distribution.
Collecting all results then leads to the following iterative EB updates
\small 
\begin{align}
\hat{k}^{(q+1)} & =\frac{\sum_{m=1}^{n_{mc}}\sum_{t=1}^{K}L_{tm}^{(q)}}{3\sqrt{\sum_{m=1}^{n_{mc}}\sum_{t=1}^{K}\sum_{l=1}^{L_{tm}^{(q)}}\left(\mu_{ltm}^{(q)}\right)^{2}}\sqrt{K}},\label{eq:Final EB k}\\
\boldsymbol{\hat{S}}^{(q+1)} & =\left(b_{1}^{(q)}/B^{(q)},\ldots,b_{p}^{(q)}/B^{(q)}\right),\label{eq:Final EB variables}\\
\left(\hat{\nu}^{(q+1)},\hat{\lambda}^{(q+1)}\right) & =\underset{\nu,\lambda}{\arg\max}\sum_{m=1}^{n_{mc}}\log\mathcal{IG}\left(\sigma_{m}^{2(q)};\frac{\nu}{2},\frac{\nu\lambda}{2}\right),\label{eq:Final EB error variance}\\
\left(\hat{\alpha}^{(q+1)},\hat{\beta}^{(q+1)}\right) & =\underset{\alpha,\beta}{\arg\max}\sum_{m=1}^{n_{mc}}\sum_{t=1}^{K}\left[\sum_{z=1}^{Z_{tm}^{(q)}}\log\left(\alpha\left(1+d_{ztm}^{(q)}\right)^{-\beta}\right)+\sum_{l=1}^{L_{tm}^{(q)}}\log\left(1-\alpha\left(1+d_{ltm}^{(q)}\right)^{-\beta}\right)\right],\label{eq:Final EB alpha beta}
\end{align}
\normalsize
with $b_{j}^{(q)}/B^{(q)}$ the estimated probability of covariate
$j$ getting selected in the splitting rules. Here, $b_{j}^{(q)}$
represents the total count of splitting rules with $j$ at iteration
$q$ for all Monte Carlo samples combined and $B^{(q)}$ represents
the total count of all splitting rules at iteration $q.$ 

Updates $\hat{k}^{(q+1)}$ and $\boldsymbol{\hat{S}}^{(q+1)}$ are
calculated analytically, whereas updates $\left(\hat{\nu}^{(q+1)},\hat{\lambda}^{(q+1)}\right)$
correspond to standard maximum likelihood estimation of an inverse
gamma distribution. Updates $\left(\hat{\alpha}^{(q+1)},\hat{\beta}^{(q+1)}\right)$ may be solved
using the base R optim function.

\section{Overfitting of EB updates}
\label{sec:Overfitting-of-EB}

Here, we illustrate that it is unfeasible to use the estimated hyperparameters
as stopping criterion for the stochastic EM algorithm. Typically,
the hyperparameters are updated using EM until they stabilize within
a given tolerance level. However, for BART, we empirically found that
this stopping criterion is unfeasible to halt the EM algorithm of
EB-coBART.

\begin{figure}[H]
\centering{}\includegraphics[bb=0bp 5bp 530bp 144bp,clip,scale=0.9]{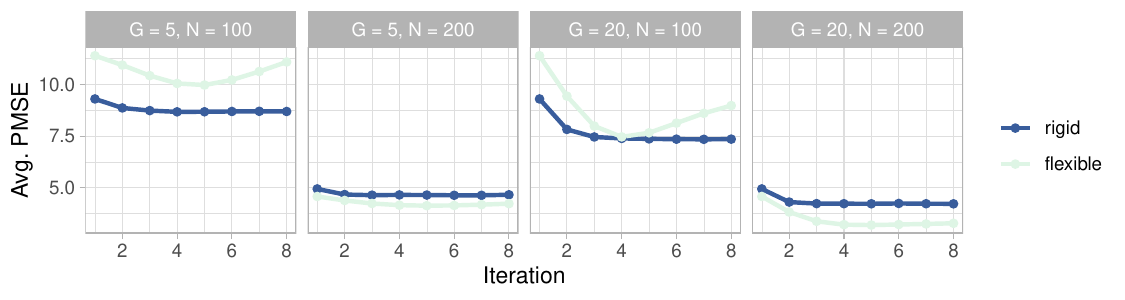}\caption{\small{Average (over data sets) prediction mean square error (PMSE) as a function of the iteration number for the rigid tree setting (blue) and the flexible tree setting (green) for the sparse and nonlinear simulation with discrete co-data. The four different
panels correspond to different simulation settings.}}
\label{fig:PMSEvsIter}
\end{figure}

\begin{figure}[H]
\centering{}\includegraphics[bb=0bp 5bp 260bp 144bp,clip]{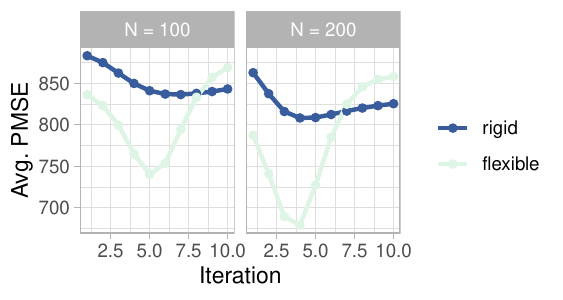}\caption{\small{Average (over data sets) prediction mean square error (PMSE) as a function of the iteration number for
the rigid tree setting (blue) and the flexible tree setting (green)
for the linear and dense simulation with continuous co-data. The left
panel corresponds to the $N=100$ simulation setting and the right
panel to the $N=200$ setting.}}
\label{fig:PMSEvsIter-LinDense}
\end{figure}

We first show overfitting of the co-data moderated EB-estimates of
$\boldsymbol{S}$ by plotting the average PMSE, with the average taken
over the $500$ simulated data sets, as a function of the iteration
number for the simulations described in the main text (Section 4). We remind the
reader that at each iteration hyperparameter $\boldsymbol{S}$ is
updated by eq. 3.12 of the main text. \autoref{fig:PMSEvsIter} shows these results for the sparse and nonlinear simulation (Subsection $4.1$ of main text) and \autoref{fig:PMSEvsIter-LinDense} shows results for the dense
and linear simulation (Subsection $4.2$ of main text and supplementary
\autoref{sec:Results_for_LinDense}). The simulation settings
are specified in the different panels of \autoref{fig:PMSEvsIter}
and \autoref{fig:PMSEvsIter-LinDense}.

\autoref{fig:PMSEvsIter} and \autoref{fig:PMSEvsIter-LinDense} illustrate that the average test performance first decreases and then increases again, which hints to overfitting. This effect is stronger for the flexible tree models than for the rigid tree models. For the rigid tree models, the average performance stabilizes as a function of the iteration number in the sparse and nonlinear simulations. 

Next, we show the co-data moderated EB estimates of $S$ at several
iteration numbers in the sparse and nonlinear setting. Specifically,
we simulate $N_{sim}=100$ data sets according to: $y_{i}=f\left(\boldsymbol{x}_{i}\right)+\epsilon,\quad\textrm{with}\quad\epsilon\sim\mathcal{N}\left(0,1\right),\quad\textrm{for}\quad i=1,\ldots,N,$
and with
\small
\begin{equation}
f\left(\boldsymbol{x}_{i}\right)=10\sin\left(\pi x_{i1}x_{i2}\right)+20\left(x_{i101}-0.5\right)^{2}+10x_{i3}+10x_{i102},\label{eq:sparse nonlinear function-1-1}
\end{equation}
\normalsize
and $X_{ij}\sim\textrm{Unif}\left(0,1\right),$ for $j=1,\dots,p$
and $p=500.$ Thus, covariates $\left\{ 1,2,3,101,102\right\} $ are
predictive for the response and the remaining $495$ covariates are
noise. 

Co-data is defined as a grouping structure with $G=5$ groups. We
set equal-sized groups of size $100.$ We then assign covariates $1,2,\ldots,100$
to group $1$, covariates $101,\ldots,200$ to group $2$, etcetera.
Group $G$ consists of covariates $401,\ldots,500$. This distribution
of covariates among the groups ensures that predictive covariates
$\{1,2,3\}$ are in Group $1$ and that predictive covariates $\{101,102\}$
are always in Group $2$.

We use the following settings for the BART model. The tree hyperparameters are set to $\left(\alpha=0.95,\beta=2,k=2\right),$ and we fix the number of trees to $K=50.$ For the error variance hyperparameters, we set $\nu=10$ and we set $\lambda$ such that the $75$\% quantile of the prior equals $2/3\hat{\textrm{Var}}\left(\boldsymbol{y}\right).$
For the Gibbs samples, we employ $10$ independent chains each containing $10000$ samples with a burnin-period of $2000$ samples.

\begin{figure}[H]
\centering{}%
\begin{tabular}{cc}
\includegraphics[bb=3bp 14bp 504bp 230bp,clip,scale=0.5]{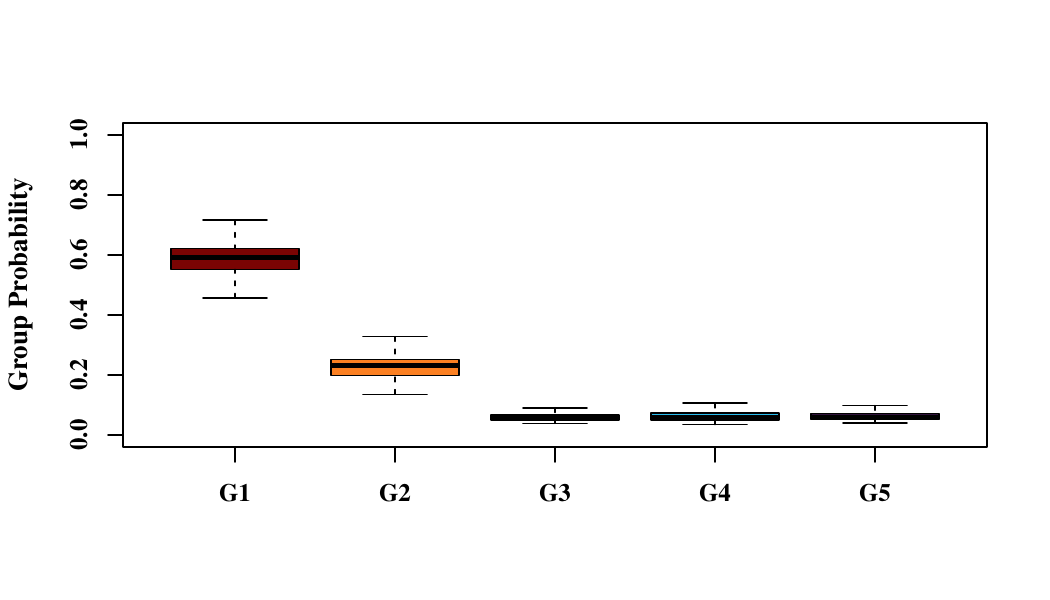} & \includegraphics[bb=3bp 14bp 504bp 230bp,clip,scale=0.5]{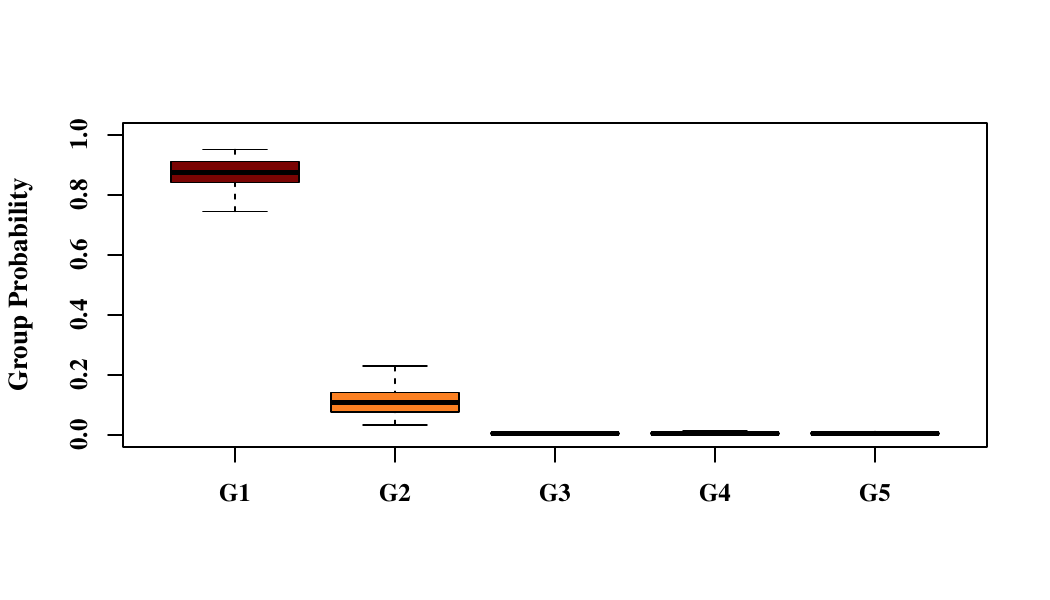}\tabularnewline
\textbf{(a)}\small{Iteration $5$}  & \textbf{(b) }\small{Iteration $7$}\tabularnewline
 & \tabularnewline
\includegraphics[bb=3bp 14bp 504bp 230bp,clip,scale=0.5]{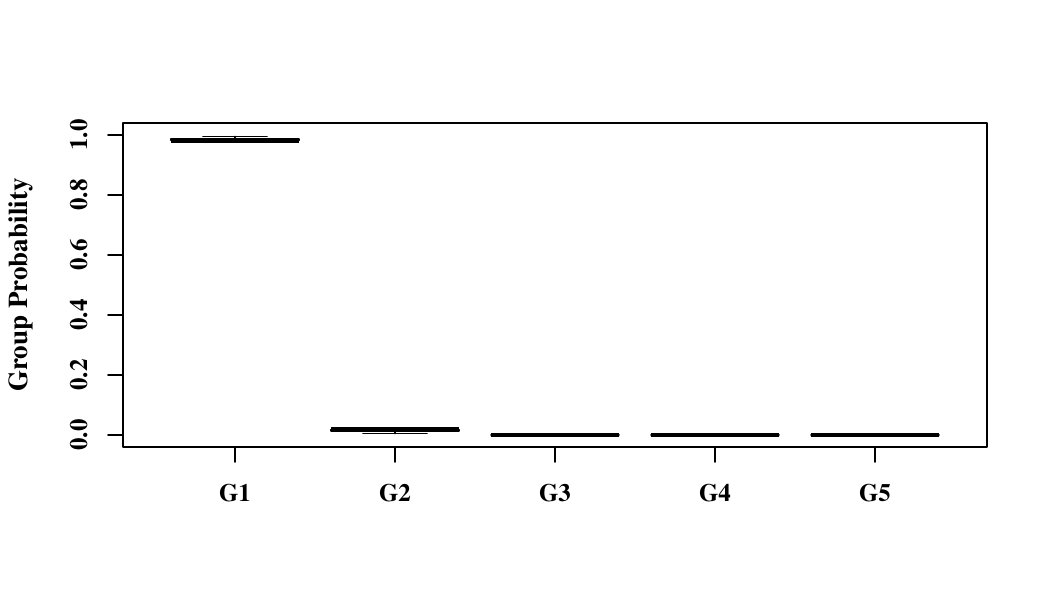} & \includegraphics[bb=3bp 14bp 504bp 230bp,clip,scale=0.5]{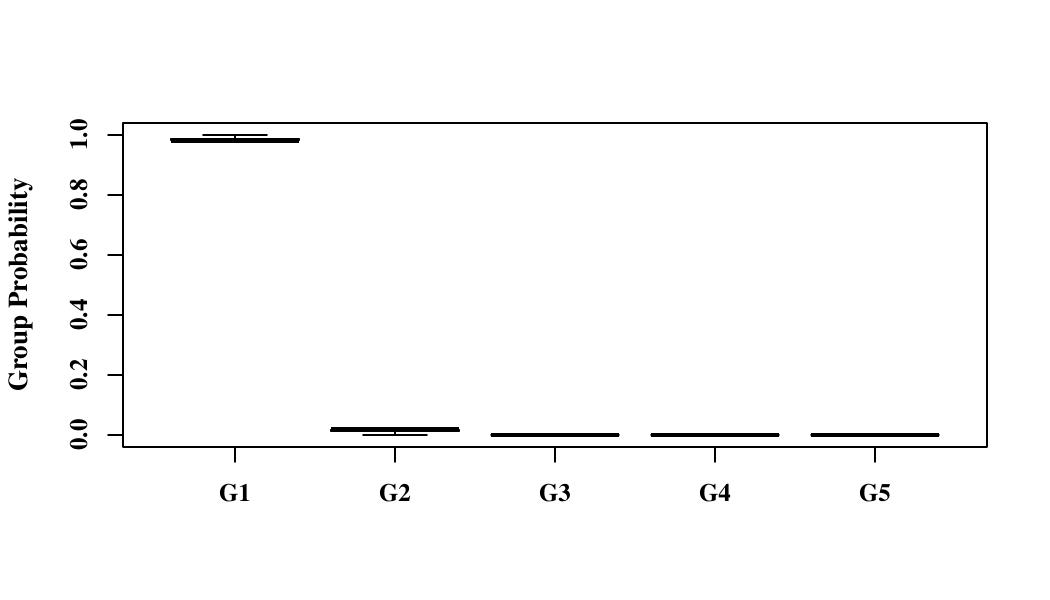}\tabularnewline
\textbf{(c)} \small{Iteration $15$} & \textbf{(d) }\small{Iteration $40$}\tabularnewline
\end{tabular}\caption{\small{Boxplots of co-data guided EB estimates
of the covariate weights across data sets at iteration $5$ \textbf{(a)},
$7$ \textbf{(b)}, $15$ \textbf{(c)}, and $40$ \textbf{(d)}.}}
\label{fig:Overfitting of EB}
\end{figure}

For each of the $100$ simulated data sets, we run EB-coBART for $40$
iterations while tracking the co-data-guided EB estimates of the covariate
weights. We show these estimated weights for several iterations in
\autoref{fig:Overfitting of EB} for all simulated data set by
depicting boxplots.

\autoref{fig:Overfitting of EB} illustrates that the estimated
group-specific covariate weights converge to the case where almost
all probability mass is put on the first group (reached maximally
at iteration $15$ for all simulated data sets). However, the second
group also has predictive covariates and hence it is undesirable to
track the co-data-guided EB estimates until these stabilize.

\section{Comparing WAIC with cross-validation}
\label{sec:Comparing-WAIC-with}

Here, we evaluate whether using the WAIC as stopping criterion for
the co-data moderated EB updates leads to similar results compared
to cross-validation as stopping criterion. To do so, we simulate $N_{sim}=100$
data sets according to: $Y_{i}=f_{1}\left(\boldsymbol{X}_{i}\right)+\epsilon,\quad\textrm{with}\quad\epsilon\sim\mathcal{N}\left(0,1\right),\quad\textrm{for}\quad i=1,\ldots,N,$
and with
\begin{equation}
f_{1}=10\sin\left(\pi X_{i1}X_{i2}\right)+20\left(X_{i101}-0.5\right)^{2}+10X_{i3}+10X_{i102},\label{eq:sparse nonlinear function-1}
\end{equation}
and $X_{ij}\sim\textrm{Unif}\left(0,1\right),$ for $j=1,\dots,p$
and $p=500.$ Thus, covariates $\left\{ 1,2,3,101,102\right\} $ are
predictive for the response and the remaining $495$ covariates are
noise.

Co-data is defined as a grouping structure with $G=5$ groups. We
set equal-sized groups of size $100.$ We then assign covariates $1,2,\ldots,100$
to group $1$, covariates $101,\ldots,200$ to group $2$, etcetera.
Group $G$ consists of covariates $401,\ldots,500$. This distribution
of covariates among the groups ensures that predictive covariates
$\{1,2,3\}$ are in Group $1$ and that predictive covariates $\{101,102\}$
are always in Group $2$.

For each simulated data set, we run EB-coBART while tracking the WAIC
(eq. 3.13 of main text), and the cross-validated prediction mean square
error (PMSE) utilizing $5$ folds. We also compute, at each iteration,
the test performance evaluated for $N_{test}=500$ independent test
samples.

We use the following settings for the BART model. The tree hyperparameters are set to $\left(\alpha=0.95,\beta=2,k=2\right),$ and we fix the number of trees to $K=50.$ For the error variance hyperparameters, we set $\nu=10$ and we set $\lambda$ such that the $75$\% quantile of the prior equals $2/3\hat{\textrm{Var}}\left(\boldsymbol{y}\right).$
For the Gibbs samples, we employ $10$ independent chains each containing $10000$ samples with a burnin-period of $2000$ samples.

\begin{figure}[H]
\centering{}
\subfloat[\small{Test performance ratio}]{
\includegraphics[bb=3bp 0bp 350bp 230bp,clip,scale=0.5]{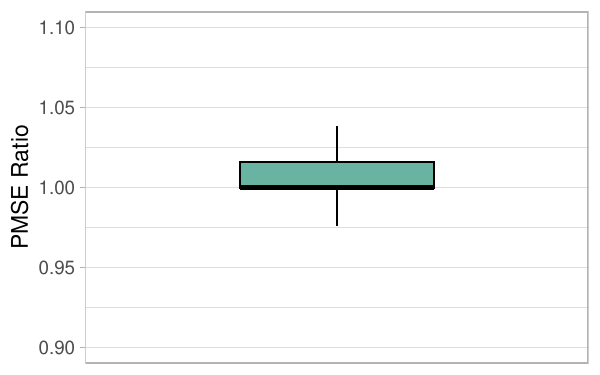}}
\subfloat[\small{Difference in estimated group-specific weights}]{
\includegraphics[bb=3bp 0bp 504bp 230bp,clip,scale=0.5]{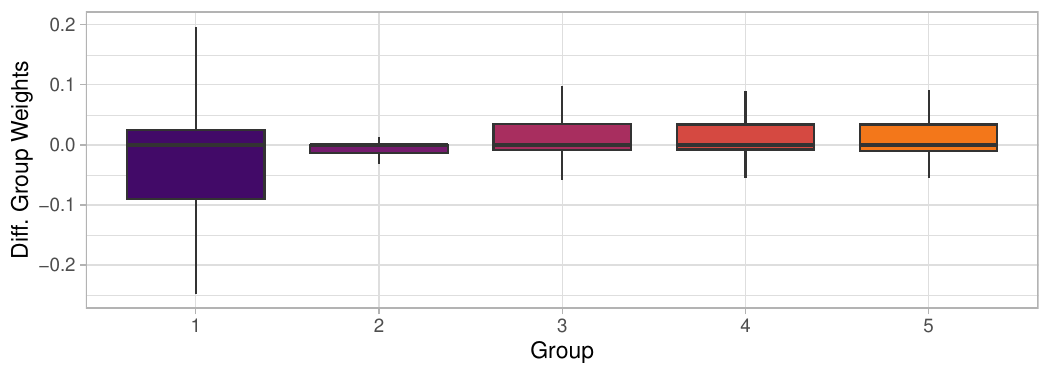}}
\caption{\small{Comparison between WAIC and cross-validation
as stopping criterion. \autoref{fig:Comparison-between-WAIC}a
depicts a boxplot of the test performance ratios across the simulated
data sets between EB-coBART with the WAIC and EBcoBART with cross-validation
as stopping criterion ($\textrm{PMSE}_{\textrm{WAIC}}/\textrm{PMSE}_{\textrm{CV}}$).
\autoref{fig:Comparison-between-WAIC}b shows boxplots of the difference
in estimated group probabilities across the data sets between EBcoBART
with the WAIC and EBcoBART with cross-validation.}}
\label{fig:Comparison-between-WAIC}
\end{figure}

To assess whether WAIC and the cross-validated test performance are
comparable as stopping criterion, we compute the ratio of test PMSE
at minimum WAIC and at minimum test CV, i.e. $\textrm{PMSE}_{\textrm{WAIC}}/\textrm{PMSE}_{\textrm{CV}},$
for each of the $100$ simulated data sets (\autoref{fig:Comparison-between-WAIC}a).
Additionally, we evaluate the difference in estimated group-specific
probability between the WAIC and CV as stopping criterion (\autoref{fig:Comparison-between-WAIC}b). 

The boxplot of the ratio of the test PMSE at minimum WAIC compared
to minimum CV, i.e. $\textrm{PMSE}_{\textrm{WAIC}}/\textrm{PMSE}_{\textrm{CV}},$
reveals that the predictive performance of the criteria is similar
with an average ratio of $1.007$ and the difference not larger than
$3\%.$ In addition, the difference in estimated group-specific weights
does not differ drastically, although for the first group, the difference
becomes larger than $0.2$ for $13$ cases. For roughly half of the
cases, the estimated group-specific weights are equal.

\section{Variable selection for the sparse and nonlinear simulation setting}
\label{sec:VarSel}
We show variable selection results for the sparse and nonlinear simulation setting described in Section 4.1 of the main text. This simulation setting is also described in \autoref{sec:Comparing-WAIC-with}. 
To quantify variable selection, we select $p_{\textrm{sel}}$ covariates having the largest count in the splitting rules. For these $p_{\textrm{sel}}$ covariates, we then compute the proportion of predictive covariate counts with respect to the total count of all covariates used in the BART model. The predictive covariates are $j=\{1,2,3,101,102\}.$ We consider $p_{\textrm{sel}}=\{5,10,20\}.$ We compute the variable selection statistics for rigid BART, flexible BART, rigid EB-coBART, and flexible EB-coBART (\autoref{tab:VarSelNonlinearSparse}).

\begin{table}[H]
\begin{centering}
\caption{\small{Variable selection for BART and EB-coBART in the rigid tree setting ($\alpha=0.1,$ $\beta=4,$ $k=1$) and the flexible tree setting ($\alpha=0.95,$ $\beta=2,$ $k=2$). Variable selection is quantified by the proportion of times a predictive covariate is selected in the $p_{\textrm{sel}}$ highest ranked covariates.}}
\label{tab:VarSelNonlinearSparse}
\begin{tabular}{ccccc}
 & Rigid BART & Rigd EB-coBART & Flexible BART & Flexible EB-coBART\tabularnewline
\hline 
\hline 
\textbf{$\boldsymbol{p_{\textrm{sel}}}$} & \multicolumn{4}{c}{$\boldsymbol{G=5,}$ $\boldsymbol{N=100}$}\tabularnewline
\hline 
$5$ best & $0.577$ & $0.589$ & $0.091$ & $0.116$\tabularnewline
$10$ best & $0.580$ & $0.593$ & $0.093$ & $0.117$\tabularnewline
$20$ best & $0.581$ & $0.594$ & $0.094$ & $0.118$\tabularnewline
$\boldsymbol{p_{\textrm{sel}}}$ & \multicolumn{4}{c}{$\boldsymbol{G=20,}$ $\boldsymbol{N=100}$}\tabularnewline
\hline 
$5$ best & $0.577$ & $0.730$ & $0.091$ & $0.208$\tabularnewline
$10$ best & $0.580$ & $0.734$ & $0.093$ & $0.212$\tabularnewline
$20$ best & $0.581$ & $0.735$ & $0.094$ & $0.214$\tabularnewline
$\boldsymbol{p_{\textrm{sel}}}$ & \multicolumn{4}{c}{$\boldsymbol{G=5,}$\textbf{ $\boldsymbol{N=200}$}}\tabularnewline
\hline 
$5$ best & $0.818$ & $0.859$ & $0.210$ & $0.241$\tabularnewline
$10$ best & $0.818$ & $0.860$ & $0.210$ & $0.241$\tabularnewline
$20$ best & $0.818$ & $0.860$ & $0.210$ & $0.241$\tabularnewline
$\boldsymbol{p_{\textrm{sel}}}$ & \multicolumn{4}{c}{$\boldsymbol{G=20,}$ $\boldsymbol{N=200}$}\tabularnewline
\hline 
$5$ best & $0.818$ & $0.932$ & $0.210$ & $0.383$\tabularnewline
$10$ best & $0.818$ & $0.932$ & $0.210$ & $0.383$\tabularnewline
$20$ best & $0.818$ & $0.932$ & $0.210$ & $0.383$\tabularnewline
\end{tabular}
\par\end{centering}
\end{table}

EB-coBART devotes a larger portion of the weight to the predictive covariates compared to BART, i.e. a BART model having equal covariate weights $s_{j}=1/p,$ in both tree flexibility settings. This effect is stronger for $G=20$ because the co-data becomes more informative compared to $G=5.$ 

Increasing $p_{\textrm{sel}}$ does not yield a larger proportion of selected predictive covariates for $N=200.$ Apparently, selecting $p_{\textrm{sel}}=5,$ which is also the number of true predictive covariates, is enough. For $N=100,$ there is a (very) small increase in the proportion of true selected covariates.

Rigid BART and rigid EB-coBART have a much larger proportion than their flexible counterparts, as expected. As discussed in the main text, a rigid tree setting favors variable selection.

\section{Results for linear and dense simulation setting}
\label{sec:Results_for_LinDense}

Here, we compare the performance of EB-coBART to that of BART, i.e. BART with equal covariate weights $s_{j}=1/p$ for a linear covariate-response relationship in a high-dimensional simulation setting.

To specify a dense and linear simulation, we simulate the $i$th instance
of response $Y$ according to $Y_{i}=\boldsymbol{X}_{i}\boldsymbol{\theta}^{T}+\epsilon_{i},$ with $\epsilon_{i}\sim\mathcal{N}\left(0,1\right)$, for $i=1,\ldots,N$, and with $p$-dimensional covariate vector $\boldsymbol{X}_{i}$ having elements $X_{ij}\stackrel{i.i.d.}{\sim}\mathcal{N}\left(0,1\right),$
for $j=1,\dots,p$ and $p=500.$ The $p$-dimensional regression parameter
$\boldsymbol{\theta}$ is a sorted vector, in decreasing order, of
values drawn from an $\textrm{Expo}\left(1\right).$ Thus, the first
covariate ($j=1$) is most predictive for the response (largest $\theta_{j}$ value) and the last covariate ($j=500$) is least predictive. We define continuous co-data by a noised-up version of the true linear effect sizes, i.e. co-data for covariate $j$ equals $\theta_{j}+\epsilon_{j}$ with $\epsilon_{j}\sim\mathcal{N}\left(0,0.2\sigma_{\boldsymbol{\theta}}\right)$, for $j=1,\ldots,P,$ and with $\sigma_{\boldsymbol{\theta}}$ the standard deviation of $\boldsymbol{\theta}.$ We consider two sample sizes: $N=100$ and $N=200$, rendering two
simulation settings.

\begin{figure}[H]
\centering{}\includegraphics[bb=0bp 0bp 495bp 174bp,clip]{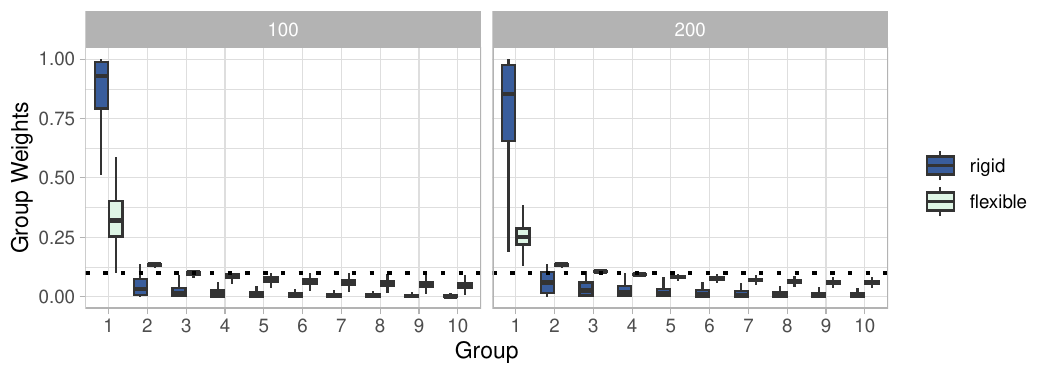}\caption{\small{Boxplots of the co-data moderated EB estimates of the covariate weights across the $500$ simulated data sets for
different simulation settings. Left (blue) boxplots correspond to EB-coBART
in the rigid tree setting ($\alpha=0.1,$ $\beta=4,$ $k=1$) and
right (green) boxplots to EB-coBART in the flexible tree setting ($\alpha=0.95,$
$\beta=2,$ $k=2$). Outliers are not shown. The horizontal dotted
lines correspond to equal group weights for $G=10.$}}
\label{fig:VarSel_LinDense}
\end{figure}

Variable importance results are shown in \autoref{fig:VarSel_LinDense}.
Because EB-coBART estimates feature specific covariate weights for
continious co-data, we visualize these estimates by first grouping
the covariates in ten groups (first $50$ covariates belonging to
group 1, the second $50$ covariates belonging to group 2, etc.) and
then reporting group averages. 

\autoref{fig:VarSel_LinDense} shows that EB-coBART upweights predictive
covariates and downweights less predictive covariates. This up/down-weighting
effect is stronger for rigid BART as this model favors variable selection
compared to flexible BART. The estimated group weights gradually decrease
because groups become less predictive for larger group number. 

\begin{figure}[h!]
\centering{}\includegraphics[bb=0bp 18bp 260bp 144bp,clip]{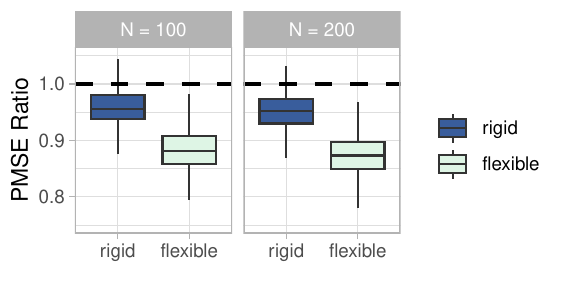}\caption{\small{Boxplot of the ratio of the prediction
mean square error (PMSE) between BART and EBcoBART ($\textrm{PMSE}_{\textrm{BART}}/\textrm{PMSE}_{\textrm{EBcoBART}}$)
across the $500$ simulated data sets for both the rigid tree setting
(left, blue) and the flexible tree setting (right, green). The two panels correspond to different simulation settings set by $N$.}}
\label{fig:RatioPMSE-LinDense}
\end{figure}

To compare the predictive performances of EB-coBART and BART,
we depict boxplots, across the $500$ simulated data sets, of the
test PMSE ratios of EB-coBART compared to BART, i.e. $\textrm{PMSE}_{\textrm{EBcoBART}}/\textrm{PMSE}_{\textrm{BART}}$
for both the rigid tree setting (left, blue) and the flexible tree setting
(right, green) (\autoref{fig:RatioPMSE-LinDense}). The left panel of
\autoref{fig:RatioPMSE-LinDense} corresponds to the $N=100$ simulation
setting, and the right panel corresponds to the $N=200$ setting.
In addition, \autoref{tab:Average-PMSE_LinDense} shows the absolute
average PMSE for EB-coBART and BART in all tree flexibility
and simulation settings. We include a comparison with ridge regression, ecpc (\cite{ecpc}), random forest, and CoRF (\cite{teBeest2017codata}).

\autoref{fig:RatioPMSE-LinDense} shows that EB-coBART has a smaller
PMSE compared to BART for most data sets (indicated by $\textrm{PMSE}_{\textrm{EB-coBART}}/\textrm{PMSE}_{\textrm{BART}}<1$).
In the rigid tree setting, BART outperforms EB-coBART for
$57$ data sets in the $N=100$ simulation setting and $40$ data
sets in the $N=200$ setting.

Inspecting the average PMSE values across the $500$ simulated data
sets reveals that on average EB-BART outperforms BART in both
tree flexiblity settings (\autoref{tab:Average-PMSE_LinDense}). Furthermore, the flexible tree models perform better than the rigid tree models. The linear prediction models ridge regression and ecpc perform much better than the BART models as expected. Ecpc strongly benefits from the co-data as indicated by the decrease in average PMSE compared to ridge regression.

Flexible EB-coBART has a lower average PMSE than random forest and its co-data extension CoRF. CoRF has a lower PMSE than random forest indicating that CoRF benefits from the co-data. Rigid EB-coBART performs worse than CoRF and slightly better than random forest.

\begin{table}[H]
\begin{centering}
\caption{\small{Average PMSE across data sets for several simulation settings for BART and EB-coBART in the rigid tree setting ($\alpha=0.1,$ $\beta=4,$ $k=1$) and the flexible tree setting ($\alpha=0.95,$ $\beta=2,$ $k=2$). Also included are four competitors: ridge regression, random forest, and co-data learners CoRF and ecpc.}}
\label{tab:Average-PMSE_LinDense}
\par\end{centering}
\centering{}%
\begin{tabular}{ccc}
 & $N=100$ & $N=200$\tabularnewline
\hline 
\hline 
Flexible BART & 836.4 & 787.5\tabularnewline
Flexible EB-coBART & 744.5 & 688.6\tabularnewline
Rigid BART & 883.4 & 863.0\tabularnewline
Rigid EB-coBART & 847.8 & 822.2\tabularnewline
DART & 836 & 786\tabularnewline
Ridge & 733.5 & 550.2\tabularnewline
Ecpc & 493.4 & 241.3\tabularnewline
Random forest & 857.5 & 826.6\tabularnewline
CoRF & 792.4 & 749.9\tabularnewline
\end{tabular}
\end{table}

\section{Uninformative co-data}
\label{sec:Uninformative-co-data}

To define an uninformative co-data setting, we simulate $N_{sim}=500$
data sets according to: $Y_{i}=f\left(\boldsymbol{X}_{i}\right)+\epsilon,\quad\textrm{with}\quad\epsilon\sim\mathcal{N}\left(0,1\right),\quad\textrm{for}\quad i=1,\ldots,N,$
and with
\begin{align}
f & =10\sin\left(\pi X_{i1}X_{i2}\right)+20\left(X_{i3}-0.5\right)^{2}+10X_{i4}+10X_{i5}\label{eq:UninformativeFunction}\\
 & +10\sin\left(\pi X_{i101}X_{i102}\right)+20\left(X_{i103}-0.5\right)^{2}+10X_{i104}+10X_{i105}\nonumber \\
 & +10\sin\left(\pi X_{i201}X_{i202}\right)+20\left(X_{i203}-0.5\right)^{2}+10X_{i204}+10X_{i205}\nonumber \\
 & +10\sin\left(\pi X_{i301}X_{i302}\right)+20\left(X_{i303}-0.5\right)^{2}+10X_{i304}+10X_{i305}\nonumber \\
 & +10\sin\left(\pi X_{i401}X_{i402}\right)+20\left(X_{i403}-0.5\right)^{2}+10X_{i404}+10X_{i505}\nonumber 
\end{align}
and $X_{ij}\sim\textrm{Unif}\left(0,1\right),$ for $j=1,\dots,p$
and $p=500.$ Sample size $N$ is fixed to $N=100$

Co-data is defined as a grouping structure with $G=5$ groups. We
set equal-sized groups of size $100.$ We then assign covariates $1,2,\ldots,100$
to group $1$, covariates $101,\ldots,200$ to group $2$, etcetera.
Group $G$ consists of covariates $401,\ldots,500$. Thus, each group
contributes equally to the overall function (\eqref{eq:UninformativeFunction})
with the firste five members being predictive and the last $95$ members
being noise. Therefore, the grouping structure is not informative
for the response.

\begin{figure}[H]
\raggedright{}\includegraphics[bb=0bp 18bp 260bp 144bp]{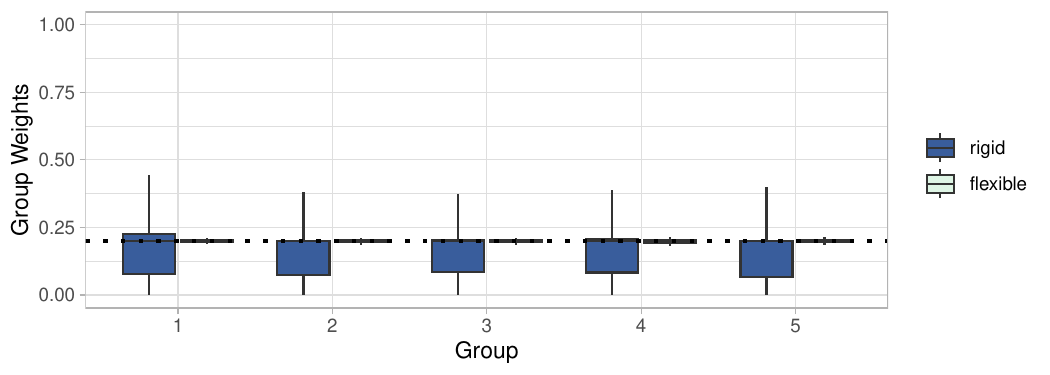}\caption{\small{Boxplots of the co-data moderated EB estimates of the covariate weights across the $500$ simulated
data sets when uninformative co-data is present. For each group, right
(blue) boxplot correspond to EB-coBART in the rigid tree setting ($\alpha=0.1,$
$\beta=4,$ $k=1$) and left (green) boxplot to EB-coBART in the flexible
tree setting ($\alpha=0.95,$ $\beta=2,$ $k=2$). Outliers are not
shown. The horizontal dotted line corresponds to equal group weights.}}
\label{fig:UninformativeCoData}
\end{figure}

\begin{figure}[H]
\centering{}\includegraphics[bb=0bp 15bp 360bp 180bp,clip]{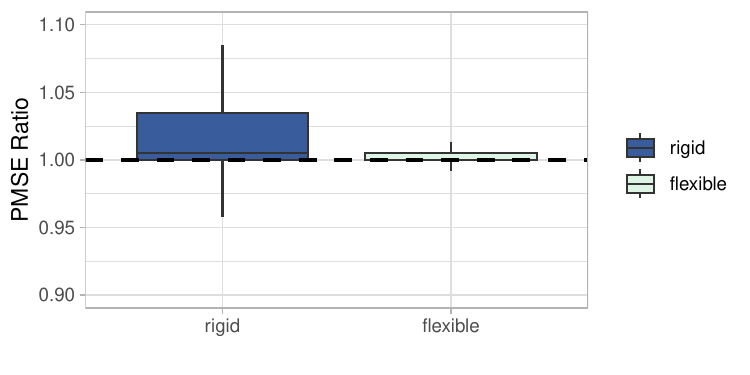}\caption{\small{Boxplot of the ratio of the prediction mean square error (PMSE) between BART and EBcoBART ($\textrm{PMSE}_{\textrm{BART}}/\textrm{PMSE}_{\textrm{EBcoBART}}$)
across the $500$ simulated data sets for both the rigid tree setting
(left, blue) and the flexible tree setting (right, green) for uninformative
co-data.}}
\label{fig:UninformativeBoxplot}
\end{figure}

We show the same results as for the simulation in the main text (Subsection
4.1). \autoref{fig:UninformativeCoData} shows boxplots across
the $500$ simulated data sets of the group-specific covariate weight
estimates $\hat{w}_{j}^{(q)},$ i.e. eq. 3.12 of the main text,
of EB-coBART having a rigid ($\alpha=0.1,$ $\beta=4,$ $k=1$) or
flexible tree setting ($\alpha=0.95,$ $\beta=2,$ $k=2$). \autoref{fig:UninformativeBoxplot} shows the ratio of test performance
of EB-coBART compared to BART, i.e. $\textrm{PMSE}_{\textrm{EBcoBART}}/\textrm{PMSE}_{\textrm{BART}}$
for both the rigid tree setting (left) and the flexible tree setting
(right) across the $500$ data sets.

The estimated group weights of the rigid tree model (left boxplots)
fluctuate around $0.2,$ i.e. equal group weights (horizontal dotted
line). The average and median (accross data sets) of each estimated
group weight equal $0.2.$ The weights of the flexible model (right
boxplots) barely fluctuate across the data sets with the average and
median also equal to $0.2$.

The fluctuations in the estimated weights induce fluctuations in the
predictive performance ratio for the rigid tree setting (left boxplot,
\autoref{fig:UninformativeBoxplot}). On average, EB-coBART and BART perform similar, indicated by an average ratio of $1.005.$
For the flexible tree setting, the PMSE ratio has a relatively constant
value of $1.00$ across the data sets. 

\section{Further results of application}
\label{sec:Further plots of application}

\begin{table}[H]
\centering{}\caption{\small{Predictive performance estimates, based on repeated ($3\times$) $10$-fold cross-validation and an external test cohort, of several prediction models.}}
\label{tab:PredictivePerfAplication}
\begin{tabular}{ccccc}
 & \multicolumn{2}{c}{\textbf{Cross-Validation}} & \multicolumn{2}{c}{\textbf{Test Set Cohort}}\tabularnewline
 & AUC & Brier score & AUC & Brier score\tabularnewline
\hline 
\hline 
Rigid BART & 0.677 & 0.166 & 0.653 & 0.160\tabularnewline
Flexible BART & 0.676 & 0.173 & 0.556 & 0.162\tabularnewline
cv-BART & 0.678 & 0.168 & 0.557 & 0.162\tabularnewline
Rigid EB-coBART 1 & 0.688 & 0.154 & 0.692 & \textbf{0.153}\tabularnewline
Rigid EB-coBART 2 & 0.713 & 0.155 & 0.703 & \textbf{0.153}\tabularnewline
Flexible EB-coBART 1 & 0.715 & 0.155 & 0.693 & 0.154\tabularnewline
Flexible EB-coBART 2 & 0.709 & 0.156 & \textbf{0.712} & \textbf{0.153}\tabularnewline
IPI-BART & 0.704 & \textbf{0.153} & 0.669 & 0.154\tabularnewline
DART & 0.70 & 0.171 & 0.552 & 0.163\tabularnewline
Random Forest & 0.688 & 0.183 & 0.637 & 0.158\tabularnewline
Corf & 0.711 & 0.157 & 0.656 & 0.159\tabularnewline
ecpc & \textbf{0.728} & 0.159 & 0.701 & 0.154\tabularnewline
Ridge & 0.723 & 0.161 & 0.689 & 0.154\tabularnewline
\end{tabular}
\end{table}

\begin{figure}[H]
\begin{centering}
\includegraphics[scale=0.9]{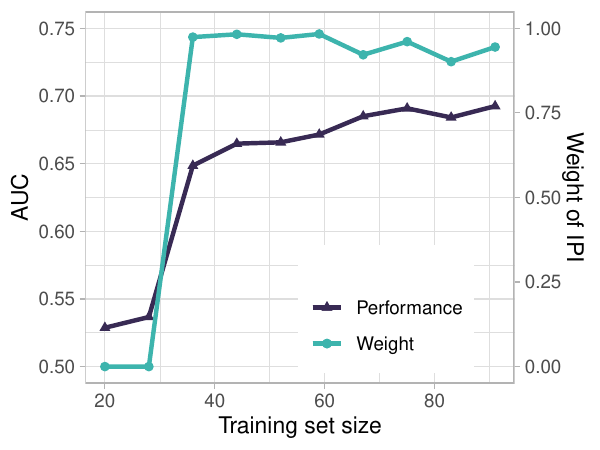}\caption{\small{Plot of the AUC (triangles, left y-axis) and the estimated weight of IPI (dots, right y-axis) as a function of the size of the training set.}}
\par\end{centering}
\label{fig:IPIandAUC_LearningCurve}
\end{figure}

\begin{figure}[H]
\begin{centering}
\includegraphics[scale=0.9]{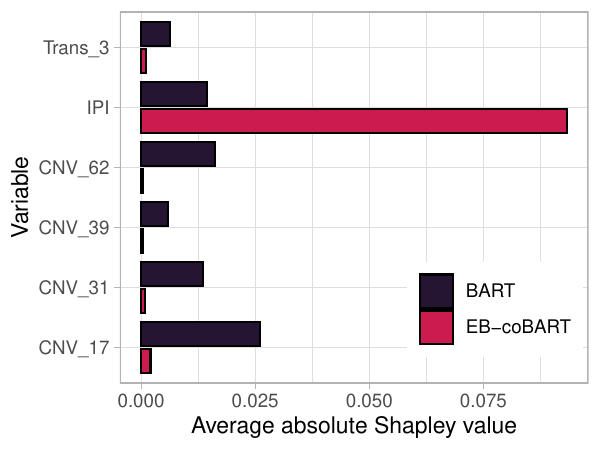}\caption{\small{Bar plot of the average absolute Shapley values for BART (black) and EB-coBART (red).}}
\par\end{centering}
\label{fig:Shapleys}
\end{figure}

Shapley values are computed using the \href{https://cran.r-project.org/web/packages/fastshap/index.html}{\texttt{fastshap}} R package \citep{ShapleyFastShap}.

\begin{figure}[H]
\subfloat[\label{fig:AlphaLearningCurve}\small{EB-estimates of $\alpha$ for several
training set sizes.}]{\raggedright{}\includegraphics[scale=0.7]{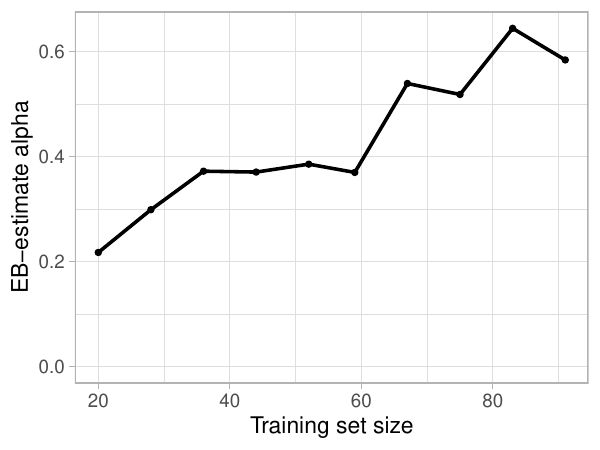}}\hspace*{\fill} \subfloat[\small{EB-estimates of $k$ for several training set sizes.}]{\raggedleft{}\label{kLearningCurve}\includegraphics[scale=0.7]{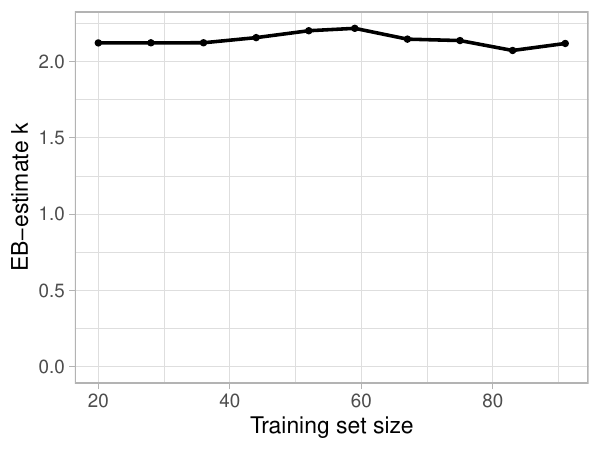}}\caption{}
\end{figure}

\section{Application 2}
\label{sec:Application 2}

This application illustrates the benefit of nonlinear co-data learners compared to linear ones. To do so, we employ the \texttt{bloodbrain} dataset from the R package \href{https://cran.r-project.org/web/packages/caret/index.html}{\texttt{caret}}. This data set has $208$ drugs for which the continuous outcome is the concentration ratio in the brain compared to the blood (log scale). The aim is to accurately predict this outcome based on  $p=134$ molecular descriptors such as charge polar surface areas \citep{Mente2005}.

We split the data set in a co-data set having $N=166$ samples, on which we estimate a co-data matrix $\boldsymbol{\mathcal{C}}$, and a primary data set having $N=42$ samples, on which we fit and evaluate the prediction models. We use a relatively small primary data set because the predictive performance of the learners saturates at moderate sample sizes ($N=60$). At $N=42,$ there is therefore some gain to be expected from the co-data, whereas for larger sample sizes the benefit of co-data will be absent for this application.

From the co-data set, we then estimate ideal co-data for our method EB-coBART, ecpc \citep{ecpc}, and CoRF \citep{teBeest2017codata} by fitting their corresponding base learners, i.e. BART, ridge regression, and random forest, respectively, and defining the continuous co-data as the estimated weights from these base learners. For EB-coBART and CoRF, these weights correspond to the count of each covariate in the tree ensemble, and for Ecpc, the weights correspond to the estimated effect sizes of the ridge regression. 

We then fit and evaluate EB-coBART, Ecpc, and CoRF on the primary data using repeated ($3\times$) $5$-fold cross-validation. We fit EB-coBART using $50$ trees and $\alpha=0.95,$ $k=2,$ and $\beta=2.$ We also considered the rigid tree setting, i.e. $\alpha=0.1,$ $k=1,$ and $\beta=4,$ but this setting performed worse. We fix the error variance, i.e. $\sigma^2,$ hyperparameters $\nu=10$ and $\lambda$ such that the $75$\% quantile of the prior equals $2/3\hat{\textrm{Var}}\left(\boldsymbol{y}\right),$ with $\hat{\textrm{Var}}\left(\boldsymbol{y}\right)$
the estimated variance of the response $\boldsymbol{y}.$ BART is fitted using $10$ independent MCMC chains each consisting of $40000$ samples of which half is burn-in. Random forest is fitted using $200$ trees. For ECPC, we also performed the post-hoc variable selection procedure explained in the main text, but this did not improve performance.

Predictive performance results, quantified by the prediction mean square error (PMSE) and $R^2$, are shown in Table . For completeness, we also show the performances of the corresponding base learners.

\begin{table}[H]
\begin{centering}
\caption{\small{Average PMSE across data sets for several simulation settings for BART and EB-coBART in the rigid tree setting ($\alpha=0.1,$ $\beta=4,$ $k=1$) and the flexible tree setting ($\alpha=0.95,$ $\beta=2,$ $k=2$). Also included are four competitors: ridge regression, random forest, and co-data learners CoRF and ecpc.}}
\label{tab:Perf_Application}
\par\end{centering}
\centering{}%
\begin{tabular}{ccc}
 & PMSE & $\textrm{R}^2$\tabularnewline
\hline 
\hline 
BART & 0.420 & 0.375\tabularnewline
EB-coBART & 0.378 & 0.438\tabularnewline
Ridge & 0.590 & 0.103\tabularnewline
Ecpc & 0.508 & 0.239\tabularnewline
Random forest & 0.404 & 0.406\tabularnewline
CoRF & \textbf{0.357} & \textbf{0.471}\tabularnewline
\end{tabular}
\end{table}

The tree-based co-data learners, which are nonlinear, have a smaller PMSE (and larger $\textrm{R}^2$) than regression-based Ecpc. CoRF performs slightly better than EB-coBART. All co-data learners show a benefit from the co-data with respect to their corresponding base-learners.

\section{Sofware and data availability}
\label{sec:Data availability and software}

R code to reproduce results is available via \url{https://github.com/JeroenGoedhart/EB_coBART_paper}.
We use R version $4.3.0.$ The repository \href{https://github.com/JeroenGoedhart/EB_coBART_paper}{EB-coBART} 
has an Application map, where the anonymized data and scripts to reproduce results presented in the application are located, and a Simulation map, which contains R scripts to reproduce the simulation results. Seeds for the pseudo random number generator are found in the scripts.

The anonymized data is a list object with $5$ elements: the train covariate data, the train response, the test covariate data, the test response, and the co-data matrix, which consist of the external information on the covariates: a grouping by covariate type and p-values on the $-\textrm{logit}$ scale.

We used the following R packages throughout the article: \href{https://cran.r-project.org/web/packages/dbarts/index.html}{dbarts (version $0.9-23$)}
to fit BART models, \href{https://cran.rstudio.com/web/packages/BART/index.html}{BART (version $2.9.6$)} to fit DART, \href{https://mc-stan.org/loo/}{loo (version $2.6.0$)}
to estimate the WAIC, \href{https://cran.r-project.org/web/packages/ecpc/index.html}{ecpc (version $3.1.1$)}
to fit ecpc and ridge regression, \href{https://www.randomforestsrc.org/}{randomForestSRC (version $3.2.2$)}
to fit random forest and CoRF, \href{https://cran.r-project.org/web/packages/pROC/index.html}{pROC (version $1.18.4$)}
to estimate the AUC, \href{https://ggplot2.tidyverse.org/}{ggplot2 (version $3.4.2$)}
to plot results.